\documentclass{article}

\usepackage{arxiv}
\usepackage{natbib}
\usepackage[utf8]{inputenc} % allow utf-8 input
\usepackage[T1]{fontenc}    % use 8-bit T1 fonts
\usepackage{hyperref}       % hyperlinks
\usepackage{url}            % simple URL typesetting
\usepackage{booktabs}       % professional-quality tables
\usepackage{amsfonts}       % blackboard math symbols
\usepackage{nicefrac}       % compact symbols for 1/2, etc.
\usepackage{microtype}      % microtypography
\usepackage{lipsum}
\usepackage{graphicx}
\graphicspath{ {./images/} }
\usepackage[nameinlink,noabbrev,capitalise,sort]{cleveref}
\usepackage{todonotes}
\usepackage{multirow}

\usepackage{pifont}%
\newcommand{\cmark}{\ding{51}}%
\newcommand{\xmark}{\ding{55}}%

\usepackage{enumitem}
\usepackage{tcolorbox}
\usepackage{subcaption}

\title{An Efficient Training Pipeline for Reasoning Graphical User Interface Agents}

\author{
 Georgios Pantazopoulos$^*$\\
  The Alan Turing Institute\\
  Heriot-Watt University\\
   \texttt{gpantazopoulos@turing.ac.uk} \\
   \And
  Eda B. \"Ozyi\u{g}it \\%\thanks{Contact: eozyigit@turing.ac.uk} \\
  The Alan Turing Institute\\
   \texttt{eozyigit@turing.ac.uk} \\
}

\begin{document}
\maketitle
\begingroup\def\thefootnote{*}\footnotetext{Work done during the internship at The Alan Turing Institute.}\endgroup
\vspace{3em}

\begin{abstract}
Visual grounding is the task of localising image regions from natural language queries and is critical for reasoning capable Graphical User Interface agents.
Many existing methods rely on massive, noisy synthetic datasets. 
This work introduces an efficient training pipeline that combines model-based data filtering with parameter-efficient fine-tuning. From 4.8M synthetic examples, 12K clean and diverse instances are curated by first identifying challenging cases, removing misaligned and then selecting a diverse set of multimodal instances. 
On this data, a 3B-parameter Vision-Language Model is trained under three regimes: supervised fine-tuning, chain-of-thought- augmented fine-tuning, and reinforcement learning via Group Relative Policy Optimization. Models trained with the filtered data and lightweight training strategies match or surpass larger baselines on benchmarks such as ScreenSpot, Multimodal-Mind2Web, and AndroidControl. 
These results demonstrate that principled data curation and robust adaptation can rival large-scale training, enabling compact yet capable multimodal reasoning agents.\footnote{The code is available at \url{https://github.com/alan-turing-institute/gui-agent}}
\end{abstract}

% keywords can be removed

\section{Introduction}

Visual grounding aims to link a natural language expression to its corresponding region in an image \cite{li2025towards, yu2016modeling, plummer2015flickr30k, pantazopoulos2025a, peng2024grounding}. This capability is foundational for emerging Graphical User Interface (GUI) agents, as advances in Large Language Models (LLMs) and Vision-Language Models (VLMs) now enable these agents to execute natural-language instructions and control desktop, mobile, and web applications \cite{zhang2025ufo2desktopagentos, wang2024mobile,zheng2024gpt4vision}. The core requirements are: multimodal perception to fuse visual and textual signals; and action execution to perform precise interface operations with mouse, keyboard, or touchscreen, commonly accompanied by lighweight planning and state tracking \cite{wu2025guiactor}. Reliable control begins with accurate grounding of instructions to specific GUI elements (e.g. buttons, icons) \citep{cheng2024seeclick, li2025screenspotpro, nayak2025uivision}.

%A necessary prerequisite for reliable control is accurate grounding of natural language instructions to the corresponding elements (e.g. buttons, icons) \citep{cheng2024seeclick, li2025screenspotpro, nayak2025uivision}.
The main approach for developing such agents equipped with exceptional grounding capabilities is to fine-tune an existing VLM on a large-scale synthetic corpora \citep{gou2025navigating, yang2024aria, lin2025showui, qin2025ui}. 
A typical pipeline first applies an off-the-shelf GUI element detector to produce candidate regions, then uses a VLM to generate referring expressions or step-level instructions tied to each region, yielding paired (screenshot, region, text) examples for supervision. This paradigm naturally leads to agents that perceive the environment exclusively through visual observations that match or surpass the performance of agents relying solely on text-based representations \citep{gou2025navigating, zhou2024webarena, gur2024a, deng2023mind2web}. However, naively fine-tuning requires large volumes of diverse data often exhibiting weak generalisation in complex and high-resolution professional settings.
With the recent success of reinforcement learning (RL) as a post-training stage, concurrent efforts \citep{luo2025gui, lu2025ui, lu2025ui, yuan2025enhancing} pivot to adopting RL by designing reward functions that promote more precise grounding and reasoning capabilities.
Group Relative Policy Optimization (GRPO) \citep{guo2025deepseek}, has recently emerged as one of the highly effective RL post-training regimes by replacing heavy value models with a simple rule-based reward system.

Furthermore, recent works \citep{ye2025limo, li2025limr} challenge the claim that reasoning necessitates excessive amounts of training data demonstrating that reasoning can emerge with only a few clean and representative examples.
On the other hand, employing synthetic pipelines for creating GUI training instances often result in noisy examples stemming from misaligned instruction-region pairs, low quality or repetitive instructions, and ambiguous regions often as a product of the detection algorithm.

\begin{figure*}[tb]
\minipage{\textwidth}
    \includegraphics[width=\linewidth]{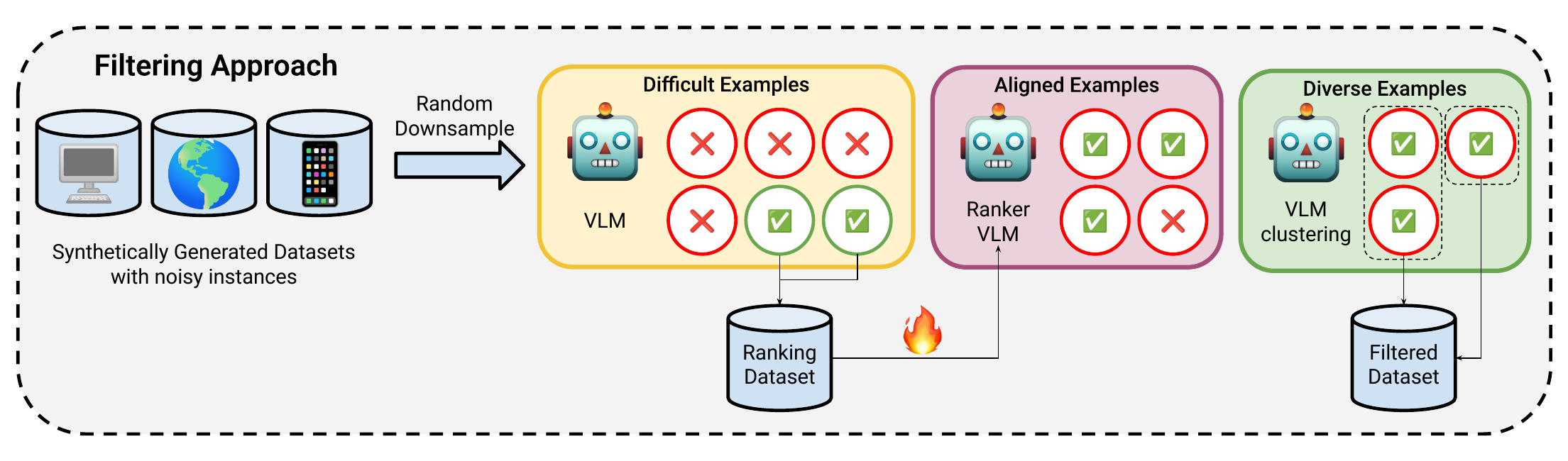}
    \label{fig:filtering_pipeline}
\endminipage\hfill
\minipage{0.49\textwidth}
  \includegraphics[width=\linewidth]{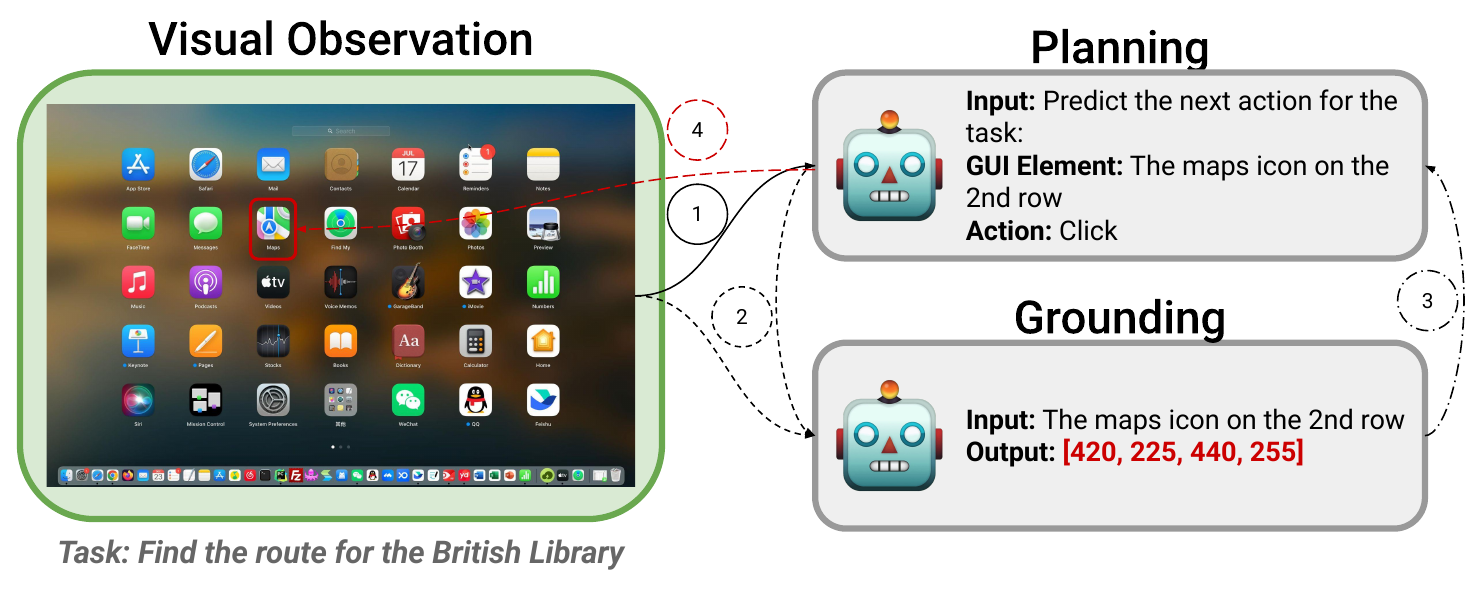}
  \label{fig:example_seeactv}
\endminipage
\minipage{0.49\textwidth}\hfill
  \includegraphics[width=0.95\linewidth]{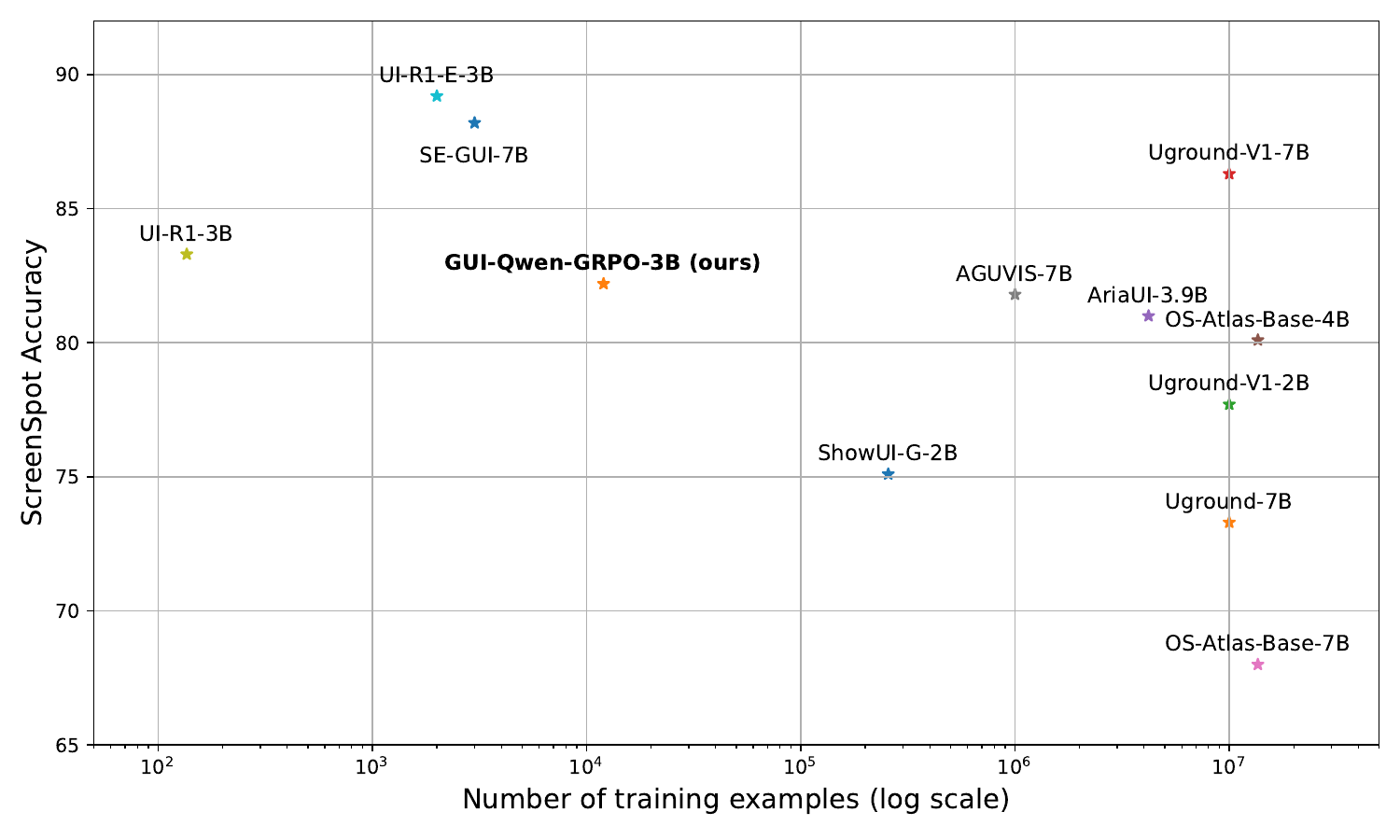}
  \label{fig:overall_results}
\endminipage
\caption{\textbf{(top)}: Overview of our filtering approach. We begin with a pool of noisy, GUI examples drawn from desktop, web, and mobile interfaces. A base VLM scores candidates, allowing us to partition the pool into easy and challenging cases. We train a ranking model on the easy subset to decide whether an instruction aligns with a candidate region in the interface. The ranker is then applied to the challenging subset to retain only aligned instances. We then cluster the embeddings from a single forward pass of the VLM and select a diverse set of challenging examples. \textbf{(bottom left)}: We integrate our grounding model into the SeeAct-V framework \citep{gou2025navigating}, which uses screenshots as the only environmental observations and performs pixel-operations via a planner VLM. \textbf{(bottom right)}: Grounding performance of ScreenSpot \citep{cheng2024seeclick}. Our model significantly outperforms prior approaches relying on supervised fine-tuning over massive amounts of data and aligns with concurrent approaches combing RL+data filtering.}
\end{figure*}

Taken together, these observations call into question the necessity of high-volume synthetic instances for developing highly-capable GUI agents. 
In this work, we examine  whether a highly-curated training seed can deliver comparable or superior performance, and explore the impact of different training regimes for building such agents.
More specifically, we begin with a collection of high-volume synthetic training examples, and using a model-based filtering approach, we derive a clean, representative, and diverse set of examples for desktop, web, and mobile applications. 
\cref{fig:filtering_pipeline} provides an overview of the stages of our filtering approach. From the initial pool of 4.8M examples, we select 12k demonstrations by determining challenging instances for the base VLM, and enforcing multimodal diversity, which yields about a compression rate of 400\%. 
We provide the chain-of-thought (CoT) traces for the intended solution from a GUI agent and explore the impact of supervised fine-tuning and GRPO with simple sparse-based rewards.
Our findings illustrate that our best-performing model matches or even surpasses the performance of similar and larger GUI agents both in terms of grounding capabilities as well as success rate in agentic evaluations. 

The main contributions of this work are: (i) a detailed filtering method for curating misaligned synthetic data stemming from data generation pipelines using the same base-model as a critic for GUI training instances; (ii) a unified framework that applies model based filtering to web-based synthetic data and training state-of-the-art VLMs that is surprisingly effective for GUI agents; and (iii) a thorough evaluation of grounding and agent capabilities in GUI interfaces (desktop, mobile, and web). The results demonstrate that our agent achieves performance, at least comparable, and often substantially better, than the state-of-the-art agents.

%\begin{itemize}[leftmargin=*]
%     \item A detailed filtering method for curating misaligned synthetic data stemming from data generation pipelines using the same base-model as a critic for GUI training instances. 
     %The outcome of our filtering approach is a clean dataset of 12k demonstrations that can be used for developing agents with modern training regimes.
%     \item A unified framework that applies model based filtering to web-based synthetic data and training state-of-the-art VLMs that is surprisingly effective for GUI agents.
%     \item A thorough evaluation of grounding and agent capabilities in GUI interfaces (desktop, mobile, and web). The results demonstrate that our agent achieves performance, at least comparable, and often substantially better, than the state-of-the-art agents.
%\end{itemize}

\section{Related Work}
\paragraph{GUI agents}
LLMs and VLMs have significantly accelerated progress in the development of GUI agents.
Early approaches relied on structured representations such as HTML or accessibility (a11y) trees \citep{zhou2024webarena, gur2024a, deng2023mind2web} for grounding with labelled bounding boxes \citep{zhang2024ufo, he2024webvoyager}.
However, text-based representations of these environments are often noisy and/or incomplete \citep{gou2025navigating} while at the same time, encoding the structure of the interface raises significant computational overhead as opposed to its corresponding visual observation. \citep{zheng2024seeact}
As a result, more recent work has shifted toward pixel-level, visually grounded agents that directly interpret screen-level observations, leveraging advances in visual encoders and multimodal reasoning to achieve robust performance in complex GUI environments \citep{cheng2024seeclick, gou2025navigating, yang2024aria, lin2025showui, hong2024cogagent, shaw2023pixels, wu2024atlas}. 
Building upon these developments, frameworks such as UI-TARS \citep{qin2025ui} and AgentS2 \citep{agashe2025agent} adopt modular or generalist-specialist architectures to improve reasoning, planning, and cross-platform generalisation \citep{bonatti2024windows, xie2024osworld}. 
Nevertheless, the predominant reliance on supervised fine-tuning (SFT) \cite{dong2024abilitiesllms} across these systems presents enduring challenges, including substantial data requirements, limited adaptability to unseen interfaces, and suboptimal scalability in dynamic GUI contexts \citep{lu2025ui, yuan2025enhancing}. 
These limitations have motivated increasing interest in learning paradigms that can leverage interaction-based or self-supervised signals to achieve more flexible and data-efficient GUI understanding.

\paragraph{Reinforcement Learning}
Reinforcement learning (RL) has recently emerged as a promising alternative to SFT for training GUI agents, offering a mechanism to optimise model behaviour through interaction-based feedback rather than static supervision \cite{zhang2025landscape}. 
Rule-based reinforcement fine-tuning (RFT) paradigms have demonstrated impressive generalisation across reasoning, code generation, and multimodal perception tasks by employing verifiable reward signals to guide model adaptation \citep{kim2025robot, wang2025pixelthink, fan2025grit}. 
Extending this paradigm to GUI environments, concurrent works such as UI-R1 \citep{lu2025ui} and GUI-R1 \citep{luo2025gui} have pioneered the integration of RFT for high-level interaction reasoning and low-level grounding, achieving improved action prediction accuracy even under limited data regimes. 
These approaches illustrate the potential of RL-based methods to enhance both the contextual understanding and decision-making efficiency of GUI agents.
Furthermore, while the aforementioned works adopt rewards tailored exclusively to the performance of the agent, similar work \citep{yuan2025enhancing} incorporates continuous reward formulations along with verifiable rewards that further guides the attention mechanisms of modern VLMs towards the target elements.
Our work is aligned with aforementioned approaches, underscoring the promise of reinforcement-driven learning as a scalable and generalisable foundation for the next generation of intelligent GUI agents, capable of bridging perception, reasoning, and control within unified multimodal frameworks.

\section{Method}
%https://docs.google.com/drawings/d/11wtww-dqCPY-zT67hf3JPsE-eIcv28nCv3KhlGZenlc/edit?usp=sharing
% \begin{figure}[tb]
%     \centering
%     \includegraphics[width=\linewidth]{figures/mainfig.pdf}
%     \caption{Overview of our filtering approach. We start with a collection of noisy and synthetically generated data from desktop, web, and mobile interfaces and determine easy and challenging examples for a VLM. We then use the easy examples to train a ranking model that determines whether an instruction-region pair is aligned in an interface. Given the ranking model we determine which of the challenging examples are aligned. Finally, we select diverse challenging examples by clustering the last hidden state of the VLM within a single forward pass.}
%     \label{fig:main_fig}
% \end{figure}
% \label{sec:methods}

\subsection{Overview}
This work follows the SeeAct-V framework \citep{gou2025navigating}, where an agent only uses screenshots as environmental observation. 
The original SeeAct-V has two stages: planning and grounding, both handled by a VLM.
For grounding, SeeAct-V uses a separate model specialised for visual grounding that directly produces the coordinates on the current screen where the agent should act.
As such a high-performing visual grounding model constitutes a necessary component for making SeeAct-V a compelling framework. 

The development of GUI models is driven by synthetic data generation pipelines where GUI annotations are created either by (i) detecting elements in the interface via pre-trained detectors \citep{lu2024omniparser} or Set-of-Marks (SOM) \citep{yang2023set} followed by a VLM that provides references for each detected element; or (ii) HTML parsing \citep{zheng2024seeact}.
Ideally, the grounding model should generalise across platforms (e.g. web, desktop, and mobile) and handle diverse ways of referring to GUI elements.
However, these pipelines often result in misaligned instances with limited variability.
To address this issue, we propose a model-based filtering approach where the base VLM is used at different stages in \cref{sec:dataset}.
Finally, \cref{sec:model} outlines our model design choices and the training regimes via supervised fine-tuning, and eliciting reasoning via chain-of-thought and reinforcement learning.

\subsection{Dataset \& Filtering Pipeline}\label{sec:dataset}
We start from a collection of existing open-source GUI grounding datasets including ShowUI-Desktop \citep{lin2025showui}, and AriaUI \citep{yang2024aria} resulting in 4.8M grounding instances covering desktop, web, and mobile interfaces. 
Although these resources are valuable, their automatic pipelines often result in noisy annotations.
To mitigate this, we employ a filtering scheme in order to select challenging, diverse, and clean examples as follows:

\paragraph{Task-Difficulty} We utilise Qwen-2.5-VL-3B \citep{bai2025qwen2} in a zero-shot setting and generate bounding box predictions for each training example.
If the prediction for an example is correct, we consider this example as easy otherwise the example is considered hard and include it in our dataset.

\paragraph{Bounding-box Accuracy} Many training resources are created using an automatic setup, either by (i) creating SOM on GUI elements and then using a VLM to create references on each element; or (ii) using text-based representations by parsing the raw  HTML or accessibility (ally) trees \citep{zhou2024webarena, deng2023mind2web}.
However, due to noise in the collection process, the hard examples obtained from the previous step may not in fact have aligned instruction-bounding box pairs.
To detect false-positive examples included in our dataset, we train a ranking model based on Qwen-2.5-VL-3B that determines whether an instruction-bounding box pair is valid. 
This ranking model is trained using the easy examples from the previous step (see \cref{sec:exp_ranker} and  \cref{sec:appendix_ranker} for further information).

\paragraph{Diverse Training Examples} An additional shortcoming of the existing resources is that the automatic data creation process results in repetitive examples.
For instance, given a GUI representing a spreadsheet, many approaches relying in a combination SOM+VLM to create annotations, yields multiple examples that point to cell entities in the GUI often with limited language variation (e.g. ``Click on Cell F42'').
To select diverse and representative training examples, we employ one forward pass and obtain the last hidden state. 
We then cluster the hidden states using PCA and k-means then obtain the examples closer to each centroid.

\paragraph{Post-Processing} Finally, we apply a post-processing filtering stage by using GPT-4o mini to further select examples where the instruction clearly points to the correct GUI element without any notion of ambiguity. 
At this stage, we also manually verified the correctness of our filtered examples.

We apply the above steps sequentially on the initial set of 4.8M examples.
With this process, we curate a clean version with 12K examples which corresponds to roughly 0.25\% of the original data.
We also note that before applying the filtering, due to computational constraints, we also randomly downsampled the Aria dataset to 10\% of its original size.

\paragraph{Generating Chain-of-thought-traces} In addition to the filtered examples, we provide chain-of-thought traces derived from GPT-4o mini. 
Given the grounding instance, we highlighted the correct region and asked the model to provide a reasoning trace for deriving the solution (see \cref{appendix:cot} for more details).

\subsection{Model}\label{sec:model} 

\paragraph{Model Architecture} In all of our experiments, we employ the instruction-tuned variant Qwen-2.5-VL-3B as our base model as it performs exceptionally well on grounding multimodal tasks and supports high-resolution image input both of which are necessary prerequisites for GUI agents.
With regards to the image resolution, typical GUI screenshots are much larger than natural images leading to significant computational overhead.
As a result, we resize images by preserving their aspect ratio to $256 \times 28 \times 28$ minimum and $1080 \times 28 \times 28$ maximum pixels, respectively.

Our goal is to explore various training regimes for efficiently developing GUI agents.
For this purpose, we resort to parameter-efficient methods for training, with low-rank adaptation (LoRA) \citep{hu2022lora} being the most prominent approach. 
% LoRA involves replacing each weight matrix $W$ from the original pre-trained model with a modified version expressed as $W^{'} = W+ B \times A$, where the matrices $B$ and $A$ contain significantly fewer parameters compared to the original weight matrix $W$.
% % By decomposing the weight updates into these low-rank matrices, LoRA effectively creates a compact, low-dimensional representation of the changes imparted during the fine-tuning process. 
% This approach dramatically reduces the number of trainable parameters while preserving model performance, making it particularly valuable for scenarios with limited computational resources. 
% Consequently, LoRA has emerged as an efficient alternative to full fine-tuning.
Within VLMs, the design choice of where to apply LoRA adapters depends on various factors.
Applying adapters on the vision encoder and/or the visual connector can be particularly effective especially when the base model has not been exposed to the target visual domain.
Conversely, language adapters can be effective when the base model is not familiar with the output text distribution.
Finally, applying adapters to all components can be effective when the base model has not been significantly exposed to the target domain.
In this work, following prior approaches \citep{laurenccon2024matters}, we explore the impact of LoRA adapters on the visual and the language backbone of a GUI-specific model. 
More specifically, we apply adapters on all linear modules and explore three variants: 1) adapters on the visual backbone and the connector; 2) adapters on the language backbone; and 3) adapters on all components of the VLM.

\paragraph{Training Regime} With regards to the training objectives, we compare three standard methods: 1) Supervised fine-tuning (SFT), where the model directly predicts the candidate element in the interface; 2) Supervised fine-tuning with chain-of-thought traces (SFT+CoT) where the model first derives a plausible explanation and then provides the candidate element; and 3) Reinforcement Learning (RL) with verifiable rewards using GRPO \citep{shao2024deepseekmath}. 
From a practical standpoint, the textual format of a candidate element usually corresponds to 10-15 tokens.
On the other hand, the context size usually is bloated with image tokens due to high-resolution demands and possibly text tokens that describe the task, output format, and the action space.
As a result, the model can only spend a limited and finite amount of steps to reason over the context.
Eliciting reasoning capabilities via prolonged generated output in the form of explanations can potentially alleviate this issue.
However, it is not straightforward to compile reasoning traces that act as a good supervision signal for the model.
For this reason, it is common to employ post-training RL by directly optimising performance via reward tailored to the downstream metric.

\paragraph{Training Details}
Throughout our experiments, we apply eight different LoRa configurations for the vision, language, and the joint model resulting in 24 different runs for each training regime.
The best candidate for each training regime is determined by the grounding performance on GUI interfaces (\cref{sec:gui_grounding}), which is then used within the SeeAct-V framework (\cref{sec:agent_eval}).
For GRPO, we formulate our reward signal based on three scalar values: 1) \textit{Format-based}, where the model receives +1 if the output follows the desired format else 0; 2) \textit{Solution-based}, where the model receives +1 if the center of the predicted bounding box falls within the ground-truth region; and 3) \textit{Length-based} as from preliminary experiments we observed that the model collapsed resulting in gibberish generations. For this purpose, we provide a +1 credit if the length of the output does not exceed 100 tokens else 0. 
The final reward is the sum of the above values without any additional scaling.
Details regarding hyperparameters are presented in \cref{sec:appendix_experiments}.

\section{Experiments}
We begin with experiments regarding the quality of our ranker (\cref{sec:exp_ranker}) showcasing the effectiveness of our model on retrieving aligned instruction-bounding box instances.
In \cref{sec:gui_grounding}, we evaluate the grounding capabilities of our model under two settings, human-generated instructions, as well as instructions generated from planners that highlight which tests for the end-to-end effectiveness of our model when integrated into the SeeAct-V framework.
Finally, we integrate our agent within the See-Act-V framework, and conduct offline evaluations (\cref{sec:agent_eval}), where our agent accepts instructions in web, mobile, and desktop applications and controls the interface of a GUI by predicting the low-level actions that satisfy a user instruction.

\subsection{Ranker model}\label{sec:exp_ranker}
The ranking  model is trained on instances where the base model correctly identifies the target element in the interface.
More specifically, given an image-instruction/expression-bbox triplet the ranker predicts whether the instruction or the expression matches the element in the image indicated by the bounding box.

\paragraph{Creating negative annotations} An important note here is to clarify how the negative image-instruction-bbox triplets are derived.
In principle, we could randomly sample a candidate bounding box, though this approach yields easy negatives. Thus, during filtering, the ranking model will be heavily biased towards the positive class.
For this reason, we include examples in the training set of the ranking model by grouping instances from the same image. 
Given an $N$ annotations from the same image, we include this image in the training set if there are at least $M$ correct predictions. 
Given that the AriaUI contains significantly more instances than ShowUI, we set $M=5$ for AriaUI and $M=1$ for ShowUI desktop, respectively.
With this process, we create 15k instances corresponding to 130k positive image-instruction-bbox triplets. 

For a given example during training, we create positive triplets by keeping the original annotations with 50\% probability. 
For negative examples, we swap the bounding box in the triplet with a different candidate belonging to the same image.
Note that in the cases where an image only contains a single annotation, this example is considered by default positive.

\begin{figure*}[tb]
    \centering
    \includegraphics[width=\linewidth]{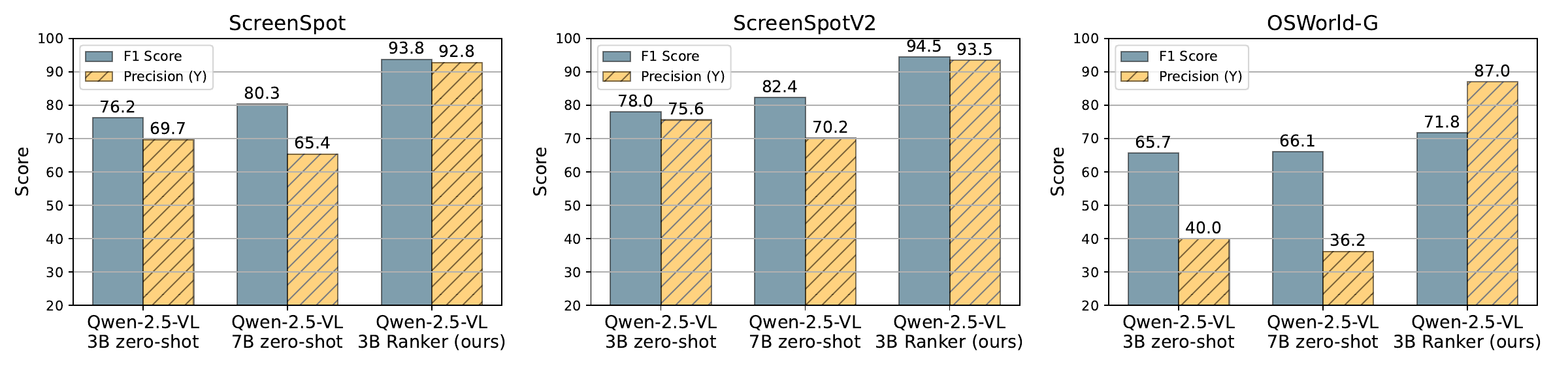}
    \caption{Performance of our ranking model against 3B/7B variants on ScreenSpot, ScreenSpotv2 and Osworld-G.}
    \label{fig:ranker_overall}
\end{figure*}

\paragraph{Results} We provide intermediate evaluations that substantiate the effectiveness of our ranking model by converting ScreenSpot \citep{cheng2024seeclick}, Screenspotv2 \citep{wu2024atlas} and Osworld-G \citep{xie2024osworld}, three popular GUI grounding benchmarks, that we converted into a binary classification, consistent with our training protocol.
Similar to the training setup, we group instances from the same image. 
If an image contains $n$ annotations, for each example $i$, we create $n-1$ negative examples by using the $i$-th instruction and all $m$ ($m \ne i$) bounding boxes from all remaining examples.

We note that in this type of evaluation 55.7\% and 71.5\% of the examples correspond to negative labels for ScreenSpot and Osworld-G, respectively.
For this reason, we primarily focus on the overall F1-score and the precision of the positive class as we are interested in measures that quantify the correctness of the filtering approach for true-positive instances.
As shown in \cref{fig:ranker_overall}, our ranking model outperforms the 3B/7B zero-shot variants, demonstrating its effectiveness within our data filtering pipeline. However, we also report overall performance metrics including F1-score, accuracy, precision, and recall, along with class-specific metrics provided in in \cref{sec:appendix_ranker}.

Finally, we note two important caveats regarding these results. 
First, the benchmarks contain clean, human-generated annotations, which likely yield an optimistic estimate of model performance on the noisy, synthetic data prevalent in training resources. 
% Training datasets frequently exhibit misaligned bounding boxes due to detection errors, overlapping screenshots that obscure interface elements, or hallucinated descriptions. 
Second, while these benchmarks serve as the best available proxy for evaluating quality-filtering models on GUI instances, the performance gap between clean benchmark and noisy training data remains uncertain. 
Despite these limitations, the results provide evidence for the viability of our ranking-based filtering approach.

\begin{table*}[tb]
    \centering
    \scriptsize
    % \small
    \renewcommand{\arraystretch}{1.2}
    \addtolength{\tabcolsep}{-0.15em}
    \begin{tabular}{@{}ll cc cc cc cc c cc cc cc cc@{}} 
        \toprule
         & \multicolumn{8}{c}{\textbf{ScreenSpot}} & & \multicolumn{8}{c}{\textbf{ScreenSpotv2}}\\
        \cline{3-10}
        \cline{12-19}
        & & \multicolumn{2}{c}{Mobile} & \multicolumn{2}{c}{Desktop} & \multicolumn{2}{c}{Web} & \multicolumn{2}{c}{Average} & & \multicolumn{2}{c}{Mobile} & \multicolumn{2}{c}{Desktop} & \multicolumn{2}{c}{Web} & \multicolumn{2}{c}{Average}\\
        \textbf{Model} & \textbf{Params} & Text & Icon & Text & Icon & Text & Icon & Micro & Macro & & Text & Icon & Text & Icon & Text & Icon & Micro & Macro\\
        \midrule
        % Fuyu & 8B & 41.0 &  1.3 &  33.0 &  3.6 &  33.9 &  4.4 & - & 19.5 & & - & - & - & - & - & - & - & -\\
        % CogAgent & 18B & 67.0 &  24.0 &  74.2 &  20.0 &  70.4 &  28.6 & 49.6 & 47.4 & & - & - & - & - & - & - & - & -\\
        % SeeClick & 9.6B & 78.0 &  52.0 &  72.2 &  30.0 &  55.7 &  32.5 & 55.8 & 53.4 & & 78.4 & 50.7 & 70.1 & 29.3 & 55.2 & 32.5 & 55.1 & 52.7 \\
        % OmniParser & - & 93.9 & 57.0 & 91.3 & 63.6 & 81.3 & 51.0 & - & 73.0 & & - & - & - & - & - & - & - & -\\
        % ShowUI-G & 2B & 92.3 & 75.5 & 76.3 & 61.1 & 81.7 & 63.6 & - & 75.1 & & - & - & - & - & - & - & - & -\\
        % UGround-V1 & 2B & 89.4 & 72.0 & 88.7 & 65.7 & 81.3 & 68.9 & - & 77.7 & & - & - & - & - & - & - & - & -\\
        % UGround-V1 & 7B & 93.0 & 79.9 & 93.8 & 76.4 & 90.9 & 84.0 & - & 86.3 & & - & - & - & - & - & - & - & -\\
        % AriaUI & 32B & 92.3 & 73.8 & 93.3 & 64.3 & 86.5 & 76.2 & 82.4 & 81.0 & & - & - & - & - & - & - & - & -\\
        % OS-Atlas-Base & 4B & 85.7 & 58.5 & 72.2 & 45.7 & 82.6 & 63.1 & 70.1 & 68.0 & & 87.2 & 59.7 & 72.7 & 46.4 & 85.9 & 63.0 & 69.2 & 71.9\\
        % OS-Atlas-Base & 7B & 93.0 & 72.9 & 91.7 & 62.9 & 90.9 & 74.3 & 82.4 & 80.1 & & 95.2 & 75.8 & 90.7 & 63.6 & 90.6 & 77.3 & 84.1 & 82.0\\
        \textcolor{gray}{Fuyu \citep{Fuyu}} & \textcolor{gray}{8B} & \textcolor{gray}{41.0} & \textcolor{gray}{1.3} & \textcolor{gray}{33.0} & \textcolor{gray}{3.6} & \textcolor{gray}{33.9} & \textcolor{gray}{4.4} & \textcolor{gray}{-} & \textcolor{gray}{19.5} & & \textcolor{gray}{-} & \textcolor{gray}{-} & \textcolor{gray}{-} & \textcolor{gray}{-} & \textcolor{gray}{-} & \textcolor{gray}{-} & \textcolor{gray}{-} & \textcolor{gray}{-}\\
        \textcolor{gray}{CogAgent \citep{hong2024cogagent}} & \textcolor{gray}{18B} & \textcolor{gray}{67.0} & \textcolor{gray}{24.0} & \textcolor{gray}{74.2} & \textcolor{gray}{20.0} & \textcolor{gray}{70.4} & \textcolor{gray}{28.6} & \textcolor{gray}{49.6} & \textcolor{gray}{47.4} & & \textcolor{gray}{-} & \textcolor{gray}{-} & \textcolor{gray}{-} & \textcolor{gray}{-} & \textcolor{gray}{-} & \textcolor{gray}{-} & \textcolor{gray}{-} & \textcolor{gray}{-}\\
        \textcolor{gray}{SeeClick \citep{cheng2024seeclick}} & \textcolor{gray}{9.6B} & \textcolor{gray}{78.0} & \textcolor{gray}{52.0} & \textcolor{gray}{72.2} & \textcolor{gray}{30.0} & \textcolor{gray}{55.7} & \textcolor{gray}{32.5} & \textcolor{gray}{55.8} & \textcolor{gray}{53.4} & & \textcolor{gray}{78.4} & \textcolor{gray}{50.7} & \textcolor{gray}{70.1} & \textcolor{gray}{29.3} & \textcolor{gray}{55.2} & \textcolor{gray}{32.5} & \textcolor{gray}{55.1} & \textcolor{gray}{52.7} \\
        \textcolor{gray}{OmniParser \citep{lu2024omniparser}} & \textcolor{gray}{-} & \textcolor{gray}{93.9} & \textcolor{gray}{57.0} & \textcolor{gray}{91.3} & \textcolor{gray}{63.6} & \textcolor{gray}{81.3} & \textcolor{gray}{51.0} & \textcolor{gray}{-} & \textcolor{gray}{73.0} & & \textcolor{gray}{-} & \textcolor{gray}{-} & \textcolor{gray}{-} & \textcolor{gray}{-} & \textcolor{gray}{-} & \textcolor{gray}{-} & \textcolor{gray}{-} & \textcolor{gray}{-}\\
        \textcolor{gray}{ShowUI-G \citep{lin2025showui}} & \textcolor{gray}{2B} & \textcolor{gray}{92.3} & \textcolor{gray}{75.5} & \textcolor{gray}{76.3} & \textcolor{gray}{61.1} & \textcolor{gray}{81.7} & \textcolor{gray}{63.6} & \textcolor{gray}{-} & \textcolor{gray}{75.1} & & \textcolor{gray}{-} & \textcolor{gray}{-} & \textcolor{gray}{-} & \textcolor{gray}{-} & \textcolor{gray}{-} & \textcolor{gray}{-} & \textcolor{gray}{-} & \textcolor{gray}{-}\\
        \textcolor{gray}{UGround-V1 \citep{gou2025navigating}} & \textcolor{gray}{2B} & \textcolor{gray}{89.4} & \textcolor{gray}{72.0} & \textcolor{gray}{88.7} & \textcolor{gray}{65.7} & \textcolor{gray}{81.3} & \textcolor{gray}{68.9} & \textcolor{gray}{-} & \textcolor{gray}{77.7} & & \textcolor{gray}{-} & \textcolor{gray}{-} & \textcolor{gray}{-} & \textcolor{gray}{-} & \textcolor{gray}{-} & \textcolor{gray}{-} & \textcolor{gray}{-} & \textcolor{gray}{-}\\
        \textcolor{gray}{UGround-V1 \citep{gou2025navigating}} & \textcolor{gray}{7B} & \textcolor{gray}{93.0} & \textcolor{gray}{79.9} & \textcolor{gray}{93.8} & \textcolor{gray}{76.4} & \textcolor{gray}{90.9} & \textcolor{gray}{84.0} & \textcolor{gray}{-} & \textcolor{gray}{86.3} & & \textcolor{gray}{-} & \textcolor{gray}{-} & \textcolor{gray}{-} & \textcolor{gray}{-} & \textcolor{gray}{-} & \textcolor{gray}{-} & \textcolor{gray}{-} & \textcolor{gray}{-}\\
        \textcolor{gray}{AriaUI \citep{yang2024aria}} & \textcolor{gray}{3.9B} & \textcolor{gray}{92.3} & \textcolor{gray}{73.8} & \textcolor{gray}{93.3} & \textcolor{gray}{64.3} & \textcolor{gray}{86.5} & \textcolor{gray}{76.2} & \textcolor{gray}{82.4} & \textcolor{gray}{81.0} & & \textcolor{gray}{-} & \textcolor{gray}{-} & \textcolor{gray}{-} & \textcolor{gray}{-} & \textcolor{gray}{-} & \textcolor{gray}{-} & \textcolor{gray}{-} & \textcolor{gray}{-}\\
        \textcolor{gray}{OS-Atlas-Base \citep{wu2024atlas}} & \textcolor{gray}{4B} & \textcolor{gray}{85.7} & \textcolor{gray}{58.5} & \textcolor{gray}{72.2} & \textcolor{gray}{45.7} & \textcolor{gray}{82.6} & \textcolor{gray}{63.1} & \textcolor{gray}{70.1} & \textcolor{gray}{68.0} & & \textcolor{gray}{87.2} & \textcolor{gray}{59.7} & \textcolor{gray}{72.7} & \textcolor{gray}{46.4} & \textcolor{gray}{85.9} & \textcolor{gray}{63.0} & \textcolor{gray}{69.2} & \textcolor{gray}{71.9}\\
        \textcolor{gray}{OS-Atlas-Base \citep{wu2024atlas}} & \textcolor{gray}{7B} & \textcolor{gray}{93.0} & \textcolor{gray}{72.9} & \textcolor{gray}{91.7} & \textcolor{gray}{62.9} & \textcolor{gray}{90.9} & \textcolor{gray}{74.3} & \textcolor{gray}{82.4} & \textcolor{gray}{80.1} & & \textcolor{gray}{95.2} & \textcolor{gray}{75.8} & \textcolor{gray}{90.7} & \textcolor{gray}{63.6} & \textcolor{gray}{90.6} & \textcolor{gray}{77.3} & \textcolor{gray}{84.1} & \textcolor{gray}{82.0}\\
        \midrule
        \multicolumn{1}{@{}l}{\textit{SFT:}} \\
        GUI-Qwen (V) & 3B & 91.9 & 67.2 & 89.7 & 57.1 & 83.9 & 62.1 & 77.0 & 75.3 & & 96.9 & 73.8 & 96.1 & 73.8 & 85.4 & 68.4 & 83.2 & 82.4\\
        GUI-Qwen (L) & 3B & 93.4 & 71.2 & 95.4 & 62.1 & 86.1 & 70.4 & 81.2 & 79.8 & & 89.1& 66.5 & 88.9 & 59.5 & 84.6 & 60.8 & 76.5 & 74.9\\
        GUI-Qwen (VL)& 3B & 95.6 & 69.9 & 93.3 & 67.9 & 85.7 & 68.4 & 81.4 & 80.1 & & 96.9 & 79.1 & 95.6 & 69.0 & 85.4 & 71.7 & 84.0 & 82.9\\
        \midrule
        \multicolumn{1}{@{}l}{\textit{SFT+CoT:}} \\
        GUI-Qwen (V) & 3B & 86.4 & 63.8 & 86.6 & 57.9 & 83.9 & 59.2 & 74.4 & 73.0 & & 89.1 & 66.5 & 88.9 & 59.5 & 84.6 & 60.8 & 76.5 & 74.9\\
        GUI-Qwen (L) & 3B & 94.5 & 75.5 & 93.3 & 62.9 & 86.5 & 67.5 & 81.6 & 80.0 & & 96.9 & 79.1 & 95.6 & 69.0 & 85.4 & 71.7 & 84.0 & 82.9\\
        GUI-Qwen (VL) & 3B & 94.5 & 71.2 & 93.8 & 69.3 & 87.0 & 65.5 & 81.4 & 80.2 & & 97.3 & 73.8 & 96.1 & 74.6 & 85.7 & 68.4 & 83.4 & 82.6\\
        \midrule 
        \multicolumn{1}{@{}l}{\textit{GRPO}} \\
        GUI-Qwen (VL) & 3B & \textbf{97.4} & \textbf{78.2} & \textbf{93.8} & \textbf{65.0} & \textbf{89.1} & \textbf{69.9} & \textbf{83.9} & \textbf{82.2} &  & \textbf{99.2} & \textbf{84.3} & \textbf{95.0} & \textbf{69.0} & \textbf{91.4} & \textbf{71.7} & \textbf{86.6} & \textbf{85.1}\\
        \bottomrule
    \end{tabular}
    \caption{Grounding accuracy on ScreenSpot and ScreenSpotv2 (Standard Setting). (V/L/VL) denotes LoRA parameters on visual, language, or both backbones.}
    \label{tab:screenspot_standard}

\end{table*}

\begin{table*}[tb]
    \centering
    \scriptsize
    % \small
    \renewcommand{\arraystretch}{1.2}
    % \addtolength{\tabcolsep}{-0.3em}
    \begin{tabular}{@{}ll cc cc cc cc@{}} 
        \toprule
        & & \multicolumn{2}{c}{Mobile} & \multicolumn{2}{c}{Desktop} & \multicolumn{2}{c}{Web} & \multicolumn{2}{c}{Average} \\
        \textbf{Model} & \textbf{Params} & Text & Icon & Text & Icon & Text & Icon & Micro & Macro\\
        \midrule
        % SeeClick \citep{cheng2024seeclick} & 9.6B & 81.0 & 59.8 & 69.6 & 33.6 & 43.9 & 26.2 & - & 52.3\\
        % UGround-V1 \citep{gou2025navigating} & 2B & 94.1 & 77.7 & 92.8 & 63.6 & 90.0 & 70.9 & - & 81.5\\
        % UGround-V1 \citep{gou2025navigating} & 7B & 94.1 & 79.9 & 93.3 & 73.6 & 89.6 & 73.3 & - & 84.0\\
        % OS-Atlas-Base \citep{wu2024atlas} & 4B & 94.1 & 73.8 & 77.8 & 47.1 & 86.5 & 65.5 & 76.8 & 74.1\\
        % OS-Atlas-Base \citep{wu2024atlas} & 7B & 93.8 & 79.9 & 90.2 & 66.4 & 92.6 & 79.1 & 85.1 & 83.7\\
        \textcolor{gray}{SeeClick \citep{cheng2024seeclick}} & \textcolor{gray}{9.6B} & \textcolor{gray}{81.0} & \textcolor{gray}{59.8} & \textcolor{gray}{69.6} & \textcolor{gray}{33.6} & \textcolor{gray}{43.9} & \textcolor{gray}{26.2} & \textcolor{gray}{-} & \textcolor{gray}{52.3}\\
        \textcolor{gray}{UGround-V1 \citep{gou2025navigating}} & \textcolor{gray}{2B} & \textcolor{gray}{94.1} & \textcolor{gray}{77.7} & \textcolor{gray}{92.8} & \textcolor{gray}{63.6} & \textcolor{gray}{90.0} & \textcolor{gray}{70.9} & \textcolor{gray}{-} & \textcolor{gray}{81.5}\\
        \textcolor{gray}{UGround-V1 \citep{gou2025navigating}} & \textcolor{gray}{7B} & \textcolor{gray}{94.1} & \textcolor{gray}{79.9} & \textcolor{gray}{93.3} & \textcolor{gray}{73.6} & \textcolor{gray}{89.6} & \textcolor{gray}{73.3} & \textcolor{gray}{-} & \textcolor{gray}{84.0}\\
        \textcolor{gray}{OS-Atlas-Base \citep{wu2024atlas}} & \textcolor{gray}{4B} & \textcolor{gray}{94.1} & \textcolor{gray}{73.8} & \textcolor{gray}{77.8} & \textcolor{gray}{47.1} & \textcolor{gray}{86.5} & \textcolor{gray}{65.5} & \textcolor{gray}{76.8} & \textcolor{gray}{74.1}\\
        \textcolor{gray}{OS-Atlas-Base \citep{wu2024atlas}} & \textcolor{gray}{7B} & \textcolor{gray}{93.8} & \textcolor{gray}{79.9} & \textcolor{gray}{90.2} & \textcolor{gray}{66.4} & \textcolor{gray}{92.6} & \textcolor{gray}{79.1} & \textcolor{gray}{85.1} & \textcolor{gray}{83.7}\\
        \midrule
        \multicolumn{1}{@{}l}{\textit{SFT:}} \\
        GUI-Qwen (V)  & 3B & 92.3 & 66.4 & 90.2 & 57.9 & 83.5 & 61.7 & 77.0 & 75.3\\
        GUI-Qwen (L)  & 3B & 94.1 & 71.2 & 94.9 & 63.6 & 87.0 & 67.5 & 81.1 & 79.7\\
        GUI-Qwen (VL) & 3B & 96.0 & 68.6 & 93.8 & 66.4 & 86.1 & 67.0 & 81.0 & 79.6\\
        \midrule
        \multicolumn{1}{@{}l}{\textit{SFT+CoT:}} \\
            GUI-Qwen (V)  & 3B & 87.6 & 65.5 & 86.6 & 58.6 & 85.2 & 61.7 & 75.6 & 74.2\\
            GUI-Qwen (L)  & 3B & 95.6 & 76.0 & 93.8 & 62.9 & 86.5 & 65.5 & 81.7 & 80.1\\
            GUI-Qwen (VL) & 3B & 92.7 & 70.7 & 93.3 & 59.3 & 83.5 & 66.5 & 79.3 & 77.7\\
        \midrule 
        \multicolumn{1}{@{}l}{\textit{GRPO}} \\
        GUI-Qwen (VL) & 3B & \textbf{97.1} & \textbf{77.7} &  \textbf{93.8} & \textbf{63.6} & \textbf{88.7} & \textbf{70.4}  & \textbf{83.6} & \textbf{81.9}\\
        \bottomrule
    \end{tabular}
    \caption{Grounding accuracy on ScreenSpot (Agent Setting) with planner-generated referring expressions. (V/L/VL) denotes LoRA parameters on visual, language, or both backbones.} 
    \label{tab:screenspot_agent}
\end{table*}

\subsection{GUI Visual Grounding}\label{sec:gui_grounding}
We begin by evaluating the grounding performance of our model on ScreenSpot \citep{cheng2024seeclick} and ScreenSpotv2 \citep{wu2024atlas}, which are specifically designed for visual grounding on GUIs. 
Both benchmarks consist of 1,272 single-step instructions and the corresponding bounding boxes of the target elements across mobile (e.g. iOS and Android), desktop (e.g. macOS and Windows), and web environments. 
The target elements encompass diverse GUI component types, including text-based elements, icons (e.g. trash can icons), and widgets (e.g. to-do lists).

Furthermore, following prior work \citep{gou2025navigating}, we focus on two settings: 1) Standard Setting, where the instructions are written by human annotations targeting a functional description of the element (e.g. simply ``close'' to refer to the ``X'' button that closes a window); and 2) Agent
Setting, where the grounding model accepts instructions from  a planning model that focus not only functional descriptions but also visual and positional
information.
To isolate the effects of instruction quality from grounding capability, we employ the instructions generated by Uground \citep{gou2025navigating}. We primarily compare our approach against several state-of-the-art models of similar size  on ScreenSpot and ScreenSpotv2.

\paragraph{Results} We report the grounding accuracy in the Standard and Agent setting in \cref{tab:screenspot_standard} and \cref{tab:screenspot_agent}, respectively.
Among the SFT variants, GUI-Qwen (VL) with LoRA parameters on both visual and language backbones performs on par or even surpass significantly larger models.
Surprisingly, the integration of chain-of-thought reasoning shows mixed results depending on the backbone configuration. 
The combination of CoT+LoRA adapters on the vision backbone results in performance degradation compared to the SFT variant.
However, GUI-Qwen (L) with SFT+CoT improves over the base SFT version, while GUI-Qwen (VL) with CoT maintains comparable performance to its SFT counterpart.
We hypothesise that this partially stems from discrepancy between the format of the trace and the outputs of the base model.
Most importantly, our GRPO-optimised GUI-Qwen (VL) model achieves state-of-the-art results across both benchmarks. 
Similar trends can be observed in the case of the Agent setting, with planner-based instructions.

\subsection{Offline Agent Evaluation}\label{sec:agent_eval}
In this section, we provide experiments regarding offline agent evaluation on  MultimodalMind2Web, Android Control, and OmniAct encompassing web, mobile, and desktop applications following their standard evaluation protocol \citep{zheng2024seeact, li2024effects, kapoor2024omniact} described below.
Importantly, we adopt the planner-generated instructions from Uground \citep{gou2025navigating} in each benchmark to decouple the effect of the quality of the planner from the downstream performance.
As such, our comparison is focused mostly with the Uground family of models.

\begin{table*}[tb]
    \centering
    \scriptsize
    % \small
    \renewcommand{\arraystretch}{1.2}
    % \addtolength{\tabcolsep}{-0.3em}
    \begin{tabular}{@{}l l l c c c c c@{}} 
        \toprule
        \textbf{Model} & \textbf{Params} & \textbf{Planner} & \textbf{Cross-Task} & \textbf{Cross-Website} & \textbf{Cross-Domain} & \textbf{Macro Average} \\
        \midrule
        \textcolor{gray}{SeeClick \citep{cheng2024seeclick}} & \textcolor{gray}{9.6B} & \textcolor{gray}{GPT-4o} & \textcolor{gray}{32.1} & \textcolor{gray}{33.1} & \textcolor{gray}{33.5} & \textcolor{gray}{32.9} \\
        \textcolor{gray}{UGround-V1 \citep{gou2025navigating}} & \textcolor{gray}{2B} & \textcolor{gray}{GPT-4o} & \textcolor{gray}{48.6} & \textcolor{gray}{47.6} & \textcolor{gray}{47.7} & \textcolor{gray}{48.0} \\
        \textcolor{gray}{UGround-V1 \citep{gou2025navigating}} & \textcolor{gray}{7B} & \textcolor{gray}{GPT-4o} & \textcolor{gray}{50.7} & \textcolor{gray}{48.1} & \textcolor{gray}{48.5} & \textcolor{gray}{49.1} \\
        \midrule
        \multicolumn{1}{@{}l}{\textit{SFT:}} \\
        GUI-Qwen (V) & 3B & GPT-4o & 48.2 & 46.5 & 45.5 & 46.7\\
        GUI-Qwen (L) & 3B & GPT-4o & 49.3 & 47.1 & 46.3 & 47.6\\
        GUI-Qwen (VL)& 3B & GPT-4o & 49.1 & 46.4 & 46.2 & 47.2\\
        \midrule
        \multicolumn{1}{@{}l}{\textit{SFT+CoT:}} \\
        GUI-Qwen (V) & 3B & GPT-4o & 47.8 & 43.2 & 45.1 & 45.4\\
        GUI-Qwen (L) & 3B & GPT-4o & 48.3 & 46.7 & 46.1 & 47.0\\
        GUI-Qwen (VL) & 3B & GPT-4o & 47.7 & 46.2 & 46.5 & 46.8\\
        \midrule 
        \multicolumn{1}{@{}l}{\textit{GRPO}} \\
        GUI-Qwen (VL) & 3B & GPT-4o & \textbf{48.9} & \textbf{47.2} & \textbf{48.2} & \textbf{48.1}\\
        \bottomrule
    \end{tabular}
    \caption{Element accuracy on Multimodal-Mind2Web. (V/L/VL) denotes LoRA parameters on visual, language, or both backbones.}
    \label{tab:mm2web_results}
\end{table*}

\paragraph{Web: MultimodalMind2Web} We use Multimodal-Mind2Web \citep{zheng2024seeact}, which extends the original Mind2Web \citep{deng2023mind2web} benchmark using multimodal information for web tasks.
These tasks are crowdsourced ensuring they represent genuine, practical needs that real users would encounter on these platforms. 
The test split consists of 1,013 tasks spanning over 100 different websites. 
Each task is defined by a high-level task instruction and a sequence of actions, with a screenshot of the webpage before each action, as the golden trajectory. We report element accuracy, i.e. accuracy of selecting the correct element, and omit operation scores because they are orthogonal to grounding comparisons.

\noindent\begin{minipage}[tb]{0.45\textwidth}
    \scriptsize
      \centering
       \begin{tabular}{@{}l l l cc@{}}
            \toprule
            \textbf{Model} & \textbf{Params} & \textbf{Planner} & \textbf{High} & \textbf{Low} \\
            % \cmidrule(lr){5-5}
            % \cline{3-4}
            \midrule
            \textcolor{gray}{SeeClick \citep{cheng2024seeclick}} & \textcolor{gray}{9.6B} & \textcolor{gray}{GPT-4o} & \textcolor{gray}{41.8} & \textcolor{gray}{52.8} \\
            \textcolor{gray}{UGround-V1\textsuperscript{\textdagger}\citep{gou2025navigating}} & \textcolor{gray}{2B} & \textcolor{gray}{GPT-4o} & \textcolor{gray}{50.0} & \textcolor{gray}{65.0} \\
            \textcolor{gray}{UGround-V1\textsuperscript{\textdagger}\citep{gou2025navigating}} & \textcolor{gray}{7B} & \textcolor{gray}{GPT-4o} & \textcolor{gray}{49.8} & \textcolor{gray}{66.2} \\
            \midrule
            \multicolumn{1}{@{}l}{\textit{SFT:}} \\
            GUI-Qwen (V) & 3B & GPT-4o & 34.2 & 60.0\\
            GUI-Qwen (L) & 3B & GPT-4o & 34.8 & 62.6\\
            GUI-Qwen (VL)& 3B & GPT-4o & 36.2 & 63.8\\
            \midrule
            \multicolumn{1}{@{}l}{\textit{SFT+CoT:}} \\
            GUI-Qwen (V) & 3B & GPT-4o & 39.0 & 61.6\\
            GUI-Qwen (L) & 3B & GPT-4o & 36.0 & 63.2\\
            GUI-Qwen (VL) & 3B & GPT-4o & 37.4 & 62.0\\
            \midrule 
            \multicolumn{1}{@{}l}{\textit{GRPO}} \\
            GUI-Qwen (VL) & 3B & GPT-4o & \textbf{44.0} & \textbf{66.0}\\
            \bottomrule
            \end{tabular}
         \captionof{table}{Step accuracy (SA) on AndroidControl over 500 random actions from the test split. Models\textsuperscript{\textdagger} are fine-tuned for this task.}
        \label{tab:androidcontrol_results}
    \end{minipage}%
    \hfill    
    \begin{minipage}[tb]{0.45\textwidth}%
    \centering
    \scriptsize
    \begin{tabular}{@{}l l l c@{}}
            \toprule
            \textbf{Model} & \textbf{Params} & \textbf{Planner} & \textbf{Action Scores} \\
            \midrule
            \textcolor{gray}{SeeClick \citep{cheng2024seeclick}} & \textcolor{gray}{9.6B} & \textcolor{gray}{GPT-4o} & \textcolor{gray}{29.6}\\
            \textcolor{gray}{UGround-V1\citep{gou2025navigating}} & \textcolor{gray}{2B} & \textcolor{gray}{GPT-4o} & \textcolor{gray}{32.9}\\
            \textcolor{gray}{UGround-V1\citep{gou2025navigating}} & \textcolor{gray}{7B} & \textcolor{gray}{GPT-4o} & \textcolor{gray}{34.0}\\
            \midrule
            \multicolumn{1}{@{}l}{\textit{SFT:}} \\
            GUI-Qwen (V) & 3B & GPT-4o & 33.8\\
            GUI-Qwen (L) & 3B & GPT-4o & 34.2\\
            GUI-Qwen (VL)& 3B & GPT-4o & 34.0\\
            \midrule
            \multicolumn{1}{@{}l}{\textit{SFT+CoT:}} \\
            GUI-Qwen (V) & 3B & GPT-4o & 33.2\\
            GUI-Qwen (L) & 3B & GPT-4o & \textbf{34.3}\\
            GUI-Qwen (VL) & 3B & GPT-4o & 34.1\\
            \midrule 
            \multicolumn{1}{@{}l}{\textit{GRPO}} \\
            GUI-Qwen (VL) & 3B & GPT-4o & 33.8\\
            \bottomrule
            \end{tabular}
         \captionof{table}{Action scores (AS) on OmniACT.}
        \label{tab:omniact_results}
\end{minipage}%

\paragraph{Mobile: Android Control} For mobile applications, we use AndroidControl \citep{li2024effects}, a dataset with 15k tasks covering 833 apps.
Each task is also densely annotated with low-level step by step instructions that correspond to an action.
Following prior work \citep{gou2025navigating, li2024effects}, we use a subset of 500 random steps from the original set and adopt two task settings: 1) high-level control where only the high-level intent is provided; and 2) low-level control, where both the high-level intent and the corresponding low-level instruction for each timestep are available.
In both settings, we report the standard step-wise accuracy where a step is considered successful only
if all the predicted actions, elements, and arguments (if applicable) are correct.

\paragraph{Desktop: OmniAct} Finally, we evaluate our model on OmniACT \citep{kapoor2024omniact}, a collection of 9,802 tasks covering 38 desktop applications and 27 websites across different platforms.
    Each of these tasks requires the agent to provide a PyAutoGUI\footnote{https://github.com/asweigart/pyautogui} script -- a sequence of actions to complete the task on a single screenshot.
We report the action score which measures how well a code snippet containing the correct action sequence can perform the task.

\paragraph{Results} Results are presented in \cref{tab:mm2web_results}, \cref{tab:androidcontrol_results}, and \cref{tab:omniact_results} for Multimodal-Mind2Web, Android Control, and Omniact, respectively.
Our GUI-Qwen models demonstrate competitive performance across all web, mobile, and desktop domains despite using only 3B parameters. 
On the Multimodal-Mind2Web benchmark, our best-performing model achieves a macro average element accuracy of 48.1\% using GRPO that is competitive with larger models.
Across training strategies, we observe that SFT yields strong baseline performance, with the language-only LoRA variant achieving 47.6\% macro average—the best among SFT approaches. Interestingly, adding chain-of-thought reasoning (SFT+CoT) does not consistently improve performance, with most variants showing slight degradation compared to their SFT counterparts. However, GRPO training demonstrates clear advantages, improving the VL model from 47.2\% (SFT) to 48.1\%.
Additionally, we observe consistency between relative categories indicating that our model generalizes well to different web domains.
Similar observations can be made in the case of AndroidControl.
The CoT training strategy shows mixed results on mobile tasks.
For high-level control, SFT+CoT with visual-only LoRA achieves 39.0\%, outperforming the best SFT approach (36.2\%). 
However, this pattern does not hold consistently across other configurations, suggesting that the benefits of explicit reasoning may be task-dependent.
The GRPO-trained GUI-Qwen (VL) achieves 44.0\% accuracy on high-level tasks and 66.0\% on low-level tasks, representing substantial improvements over SFT baselines.
Finally, in OmniACT, all of our models cluster around 33-34\% action scores, with the SFT+CoT language-only variant achieving the highest score at 34.26\% with GRPO providing minimal improvements. 
This could be explained by the quality of the planner, as planning errors often bottleneck the overall performance \citep{gou2025navigating}.

\subsection{Analysis}

\begin{figure*}[tb]
\minipage{0.49\textwidth}
    \includegraphics[width=\linewidth]{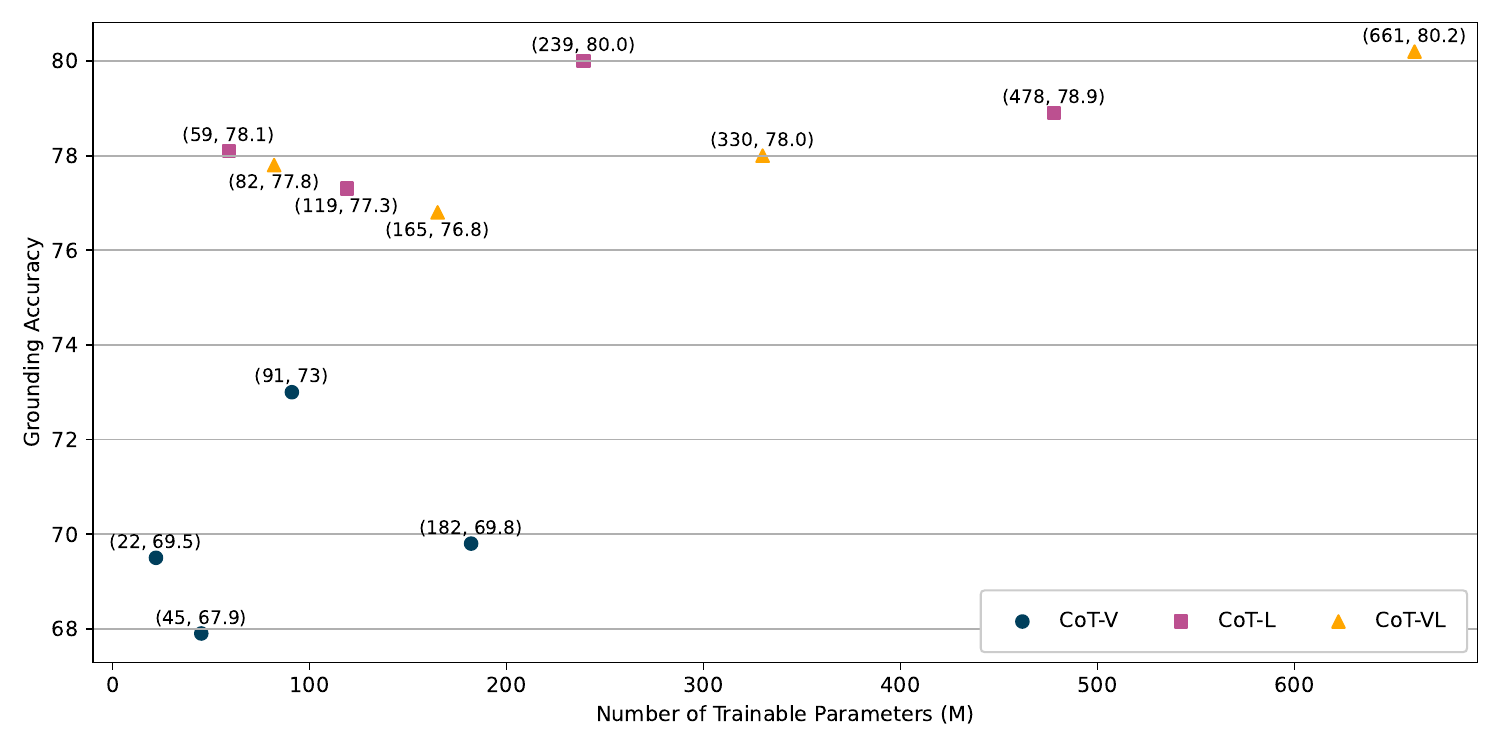}
    \subcaption{}\label{fig:cot_lora}
\endminipage\hfill
\minipage{0.49\textwidth}
  \includegraphics[width=\linewidth]{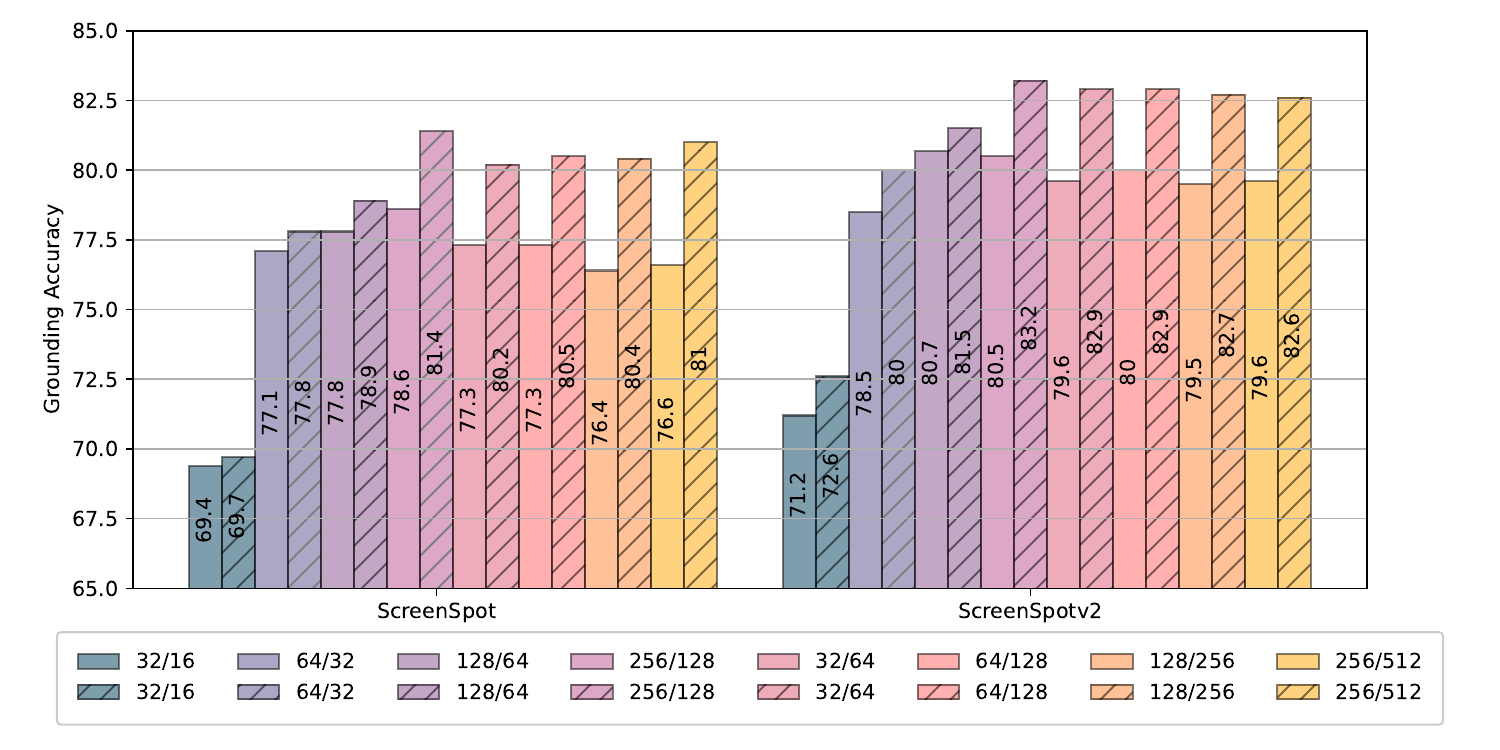}
  \subcaption{}\label{fig:data_ablations}
\endminipage
\caption{\textbf{(a)}: Performance of chain-of-thought models under different LoRA configurations. \textbf{(b)}: Data ablation results on ScreenSpot and ScreenSpotv2. We re-train all LoRA configurations (rank/alpha) for GUI-Qwen (VL) using the original data and the data from our filtering method. Bars with stripes (//) indicate training runs using our filtered data. Our filtering approach yields consistent improvements across all configurations.}
\end{figure*}

\paragraph{Which parameters of a non-thinking VLM can facilitate reasoning capabilities?}
Recall that from the previous results (e.g. \cref{tab:screenspot_standard}), applying LoRA adapters exclusively on the visual components of a VLM can yield performance degradation.
Here, we investigate this further by comparing the performance of our model under CoT training and four different LoRA adapters on the visual, language, or both backbones.
\cref{fig:cot_lora} shows the performance of these runs as a function of the number of training parameters.
We observe that by controlling the total number of training parameters, injecting LoRA adapters on the language backbone yields substantial improvement compared to the visual backbone which could be attributed to differences in the expected output text distribution.
This is in line with our previous findings as well as prior research in the field \citep{laurenccon2024matters}.
Finally, we observe moderate improvements in the case of vision-and-language over language-only adapters particularly for high-capacity configurations.

\paragraph{What type of generations are preferred by GUI reasoning models?} In the previous experiments, we built on the assumption that the reasoning trace can guide the model's bounding box prediction.
As such, we constructed succinct traces describing the interface and/or the target element.
We also observed that a model equipped with such traces enjoys benefits with regards to grounding capabilities particularly when the LoRa adapters are applied on the language backbone.
However, we also observed that models trained using GRPO exhibit substantial improvement over the chain-of-thought models without imposing any strict structure on the traces. 
Consequently, we are interested in identifying what types of traces are preferred by these models.

For this purpose, we manually reviewed 100 randomly sampled predictions from ScreenSpot using the SFT+CoT and GRPO models. We observed that both models favoured short generations, this is expected due to the short traces present during SFT+CoT training and the reward penalizing prolonged outputs during GRPO. 
However, a notable distinction is that the model trained with GRPO used the tracing tokens as a ``sketch pad'' for the input instruction. 
In the vast majority of the predictions, the model tended to repeat the entire or rephrase parts of the instruction then explicitly attempt to refer to the target element (e.g. ``play the next song'', <think>To play the next song, I should click on the right arrow icon.</think><answer>[445,1016,508,1053]</answer>''). 
This contrasts with our original hypothesis where the traces should provide a top-down summary before explicitly pointing to elements in the interface.
While this behaviour still demonstrates interpretable reasoning, it reveals a more shallow reasoning pattern focused on instruction alignment rather than the deeper visual understanding we expected from chain-of-thought prompting.

\begin{figure*}[tb]
\minipage{0.33\textwidth}
    \includegraphics[width=\linewidth]{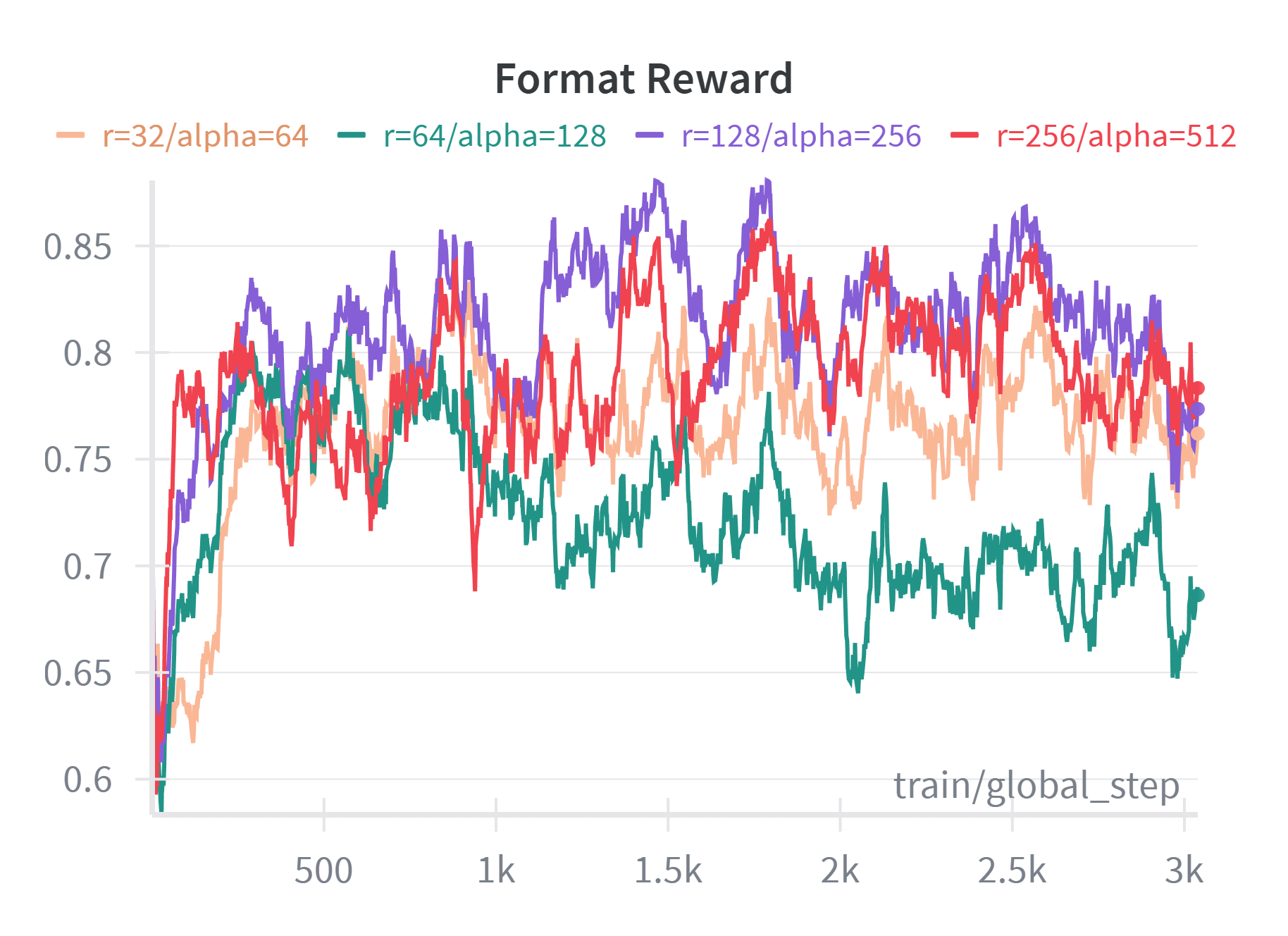}
\endminipage\hfill
\minipage{0.33\textwidth}
  \includegraphics[width=\linewidth]{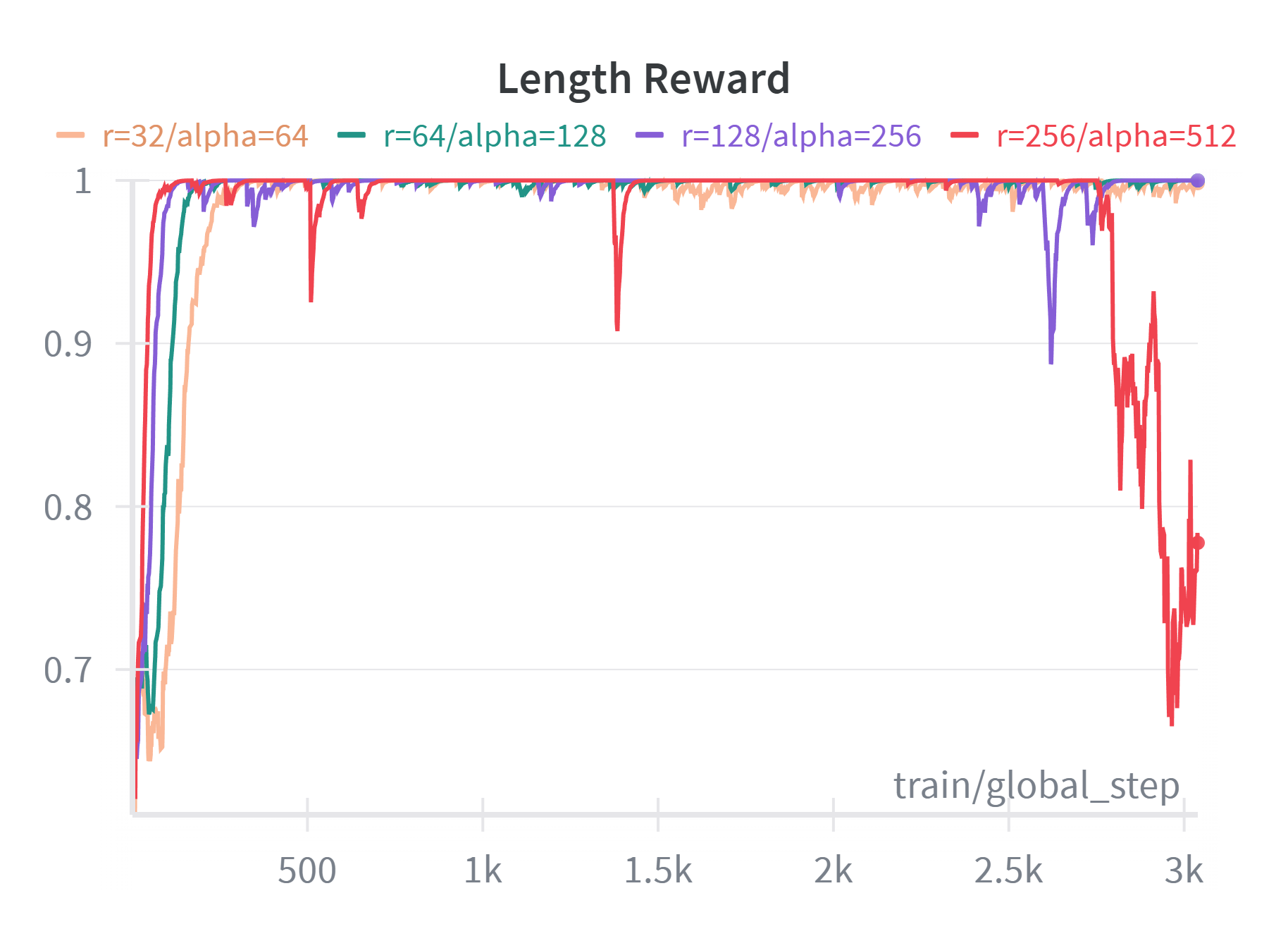}
\endminipage\hfill
\minipage{0.33\textwidth}
  \includegraphics[width=\linewidth]{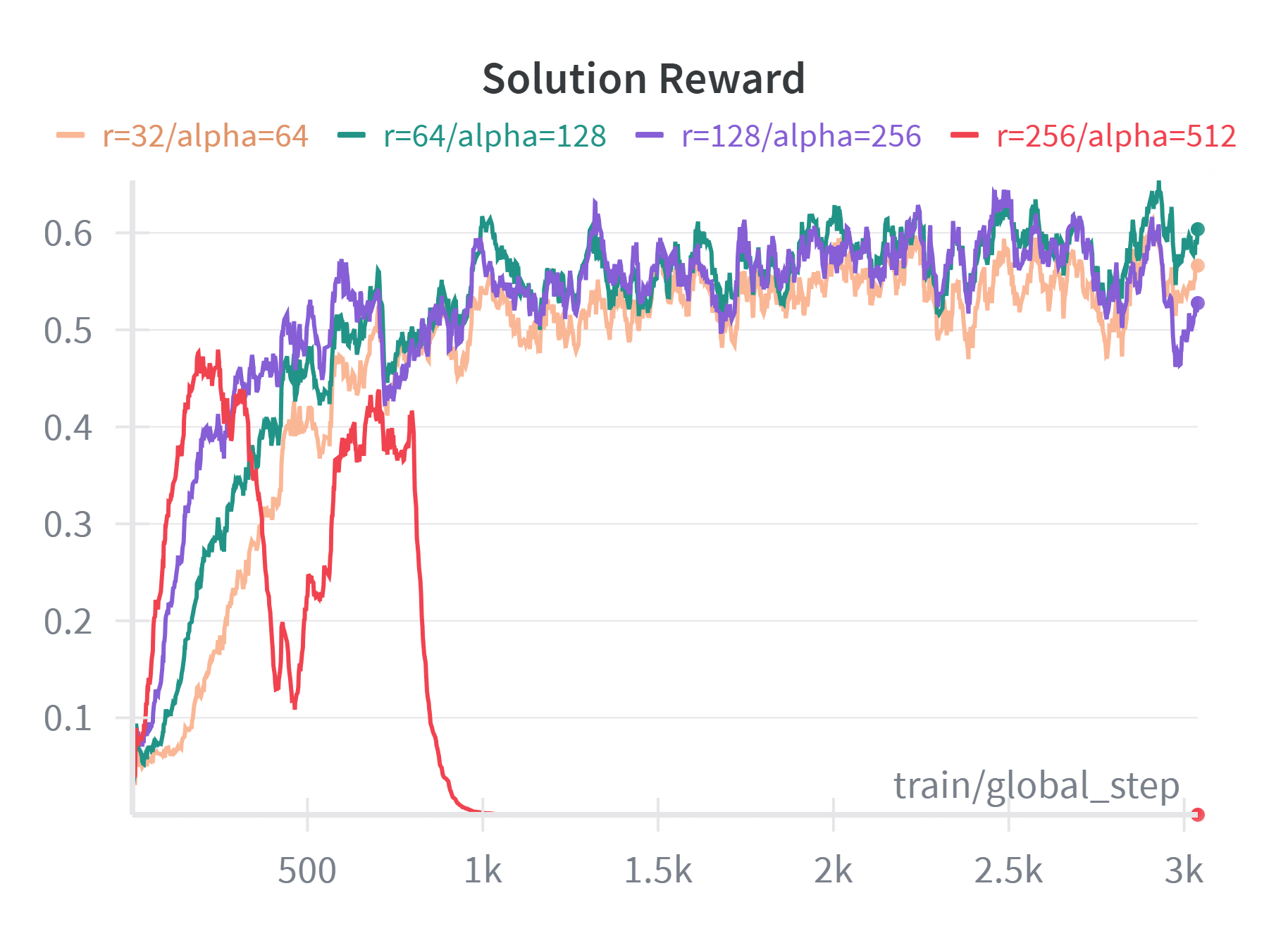}
\endminipage
\caption{Reward curves for four different LoRA configurations.}
\label{fig:rewards}
\end{figure*}

\paragraph{Sensitivity of RL} Despite outperforming both SFT and SFT/CoT regimes, GRPO training exhibits significant sensitivity to hyperparameters. \cref{fig:rewards} illustrates the reward curves for four different LoRA configurations throughout training.
While all models quickly learn to satisfy format and length constraints from the earliest training stages, we observe catastrophic model collapse in the configuration with the highest number of trainable parameters.
Specifically, this configuration initially demonstrates superior performance compared to models with fewer trainable parameters during the first 250 training steps. 
However, after approximately 500 steps, the model collapses abruptly, yielding extremely low solution rewards.
From this point until 1,000 updates, the model attempts recovery but fails to restore performance, ultimately producing zero solution rewards for the remainder of training.
We hypothesise that this collapse correlates with the interplay between the number of trainable parameters and the reward function's design.
Since format and length rewards saturate early in training, the additional model capacity may be exploited to game the solution component in pursuit of total reward maximisation.
This behaviour could potentially be mitigated by increasing the Kullback-Leibler (KL) penalty coefficient to constrain divergence from the base model, or rescaling the solution reward disproportionally to rewards related to the structure of the output.

\paragraph{Data Ablation}

% \begin{figure*}[tb]
%     \centering
%   \includegraphics[width=0.7\linewidth]{figures/ablations.pdf}
%   \caption{Data ablation results on ScreenSpot and ScreenSpotv2. We re-train all LoRA configurations for GUI-Qwen (VL) using the original data and the data from our filtering method. Bars with stripes (//) indicate training runs using our filtered data. Our filtering approach yields consistent improvements across all configurations.}
%   \label{fig:data_ablations}
% \end{figure*}

To validate the effectiveness of our approach, we conduct an ablation study with GUI-Qwen (VL) using the data before and after our filtering under supervised fine-tuning.
We facilitate a fair comparison under controlled conditions by applying the same training hyper-parameters and LoRA configurations.
\cref{fig:data_ablations} illustrates the performance of our model on ScreenSpot and ScreenSpotv2, when using the original data and the data produced by our filtering approach.
Across all settings, we observe consistent improvements when the model is trained using our own smaller but cleaner version of the original training dataset, demonstrating the value of our data curation pipeline.

% \begin{table}[tb]
%     \centering
%     \scriptsize
%     \addtolength{\tabcolsep}{-0.10em}
%     \begin{tabular}{@{}l c cc cc cc cc c cc cc cc cc@{}} 
%         \toprule
%          & \multicolumn{8}{c}{\textbf{Screenspot}} & & \multicolumn{8}{c}{\textbf{Screenspotv2}}\\
%         \cline{3-10}
%         \cline{12-19}
%          \textbf{Params} & \textbf{Filtering} &\multicolumn{2}{c}{Mobile} & \multicolumn{2}{c}{Desktop} & \multicolumn{2}{c}{Web} & \multicolumn{2}{c}{Average} & & \multicolumn{2}{c}{Mobile} & \multicolumn{2}{c}{Desktop} & \multicolumn{2}{c}{Web} & \multicolumn{2}{c}{Average}\\
%         (r/a) &  & Text & Icon & Text & Icon & Text & Icon & Micro & Macro & & Text & Icon & Text & Icon & Text & Icon & Micro & Macro\\
%         \midrule
%         \multirow{2}{*}{32/16} &  \xmark & 65.9 & 92.7 & 60.0 & 88.1 & 64.6 & 80.4 & 76.8 & 75.3 & & 72.3 & 93.0 & 65.1 & 89.4 & 64.6 & 85.0 & 79.6 & 78.2\\
%          & \cmark & 69.0 & 96.0 & 65.7 & 93.3 & 68.4 & 85.2 & 81.0 & 79.6 & & 74.9 & 96.1 & 71.4 & 95.0 & 66.7 & 86.1 & 82.6 & 81.7 \\
%         \midrule
%         \multirow{2}{*}{32/64} & \xmark & 71.6 & 90.5 & 55.0 & 88.7 & 63.6 & 80.9 & 76.8 &  75.0 & & 73.8 & 93.0 & 58.7 & 91.7 & 65.4 & 83.6 & 79.3 & 77.0\\
%         & \cmark & 69.9 & 95.6 & 67.9 & 93.3 & 68.4 & 85.7 & 81.4 & 80.1 & & 73.8 & 96.9 & 73.8 & 96.1 & 68.4 & 85.4 & 83.2 & 82.4 \\
%         \bottomrule
%         \end{tabular}
%     \label{tab:data_ablation}
%     \caption{Data ablation.}
% \end{table}

\section{Conclusion \& Limitations}
In this work, we introduced an efficient training pipeline for developing reasoning-capable GUI agents by leveraging a highly curated dataset and lightweight fine-tuning strategies. 
Our model-based filtering framework demonstrated that high-quality, diverse, and challenging examples can be distilled from large synthetic datasets, significantly reducing data volume while improving grounding accuracy. 
Through systematic comparisons of supervised fine-tuning (SFT), chain-of-thought augmented SFT (SFT+CoT), and reinforcement learning with GRPO, we showed that small-scale models can match or surpass the performance of much larger baselines. 
Our analysis indicates that parameter-efficient adaptation via LoRA adapters plays a crucial role in eliciting reasoning behaviour, particularly when applied to the language backbone. 
Additionally, the filtered dataset proved critical for both data efficiency and performance stability, underscoring the importance of data quality over sheer scale in multimodal GUI reasoning. Finally, the qualitative examination of reasoning traces highlighted that current models tend to adopt shallow, instruction-alignment reasoning patterns rather than true visual abstraction, indicating promising directions for future refinement.

\paragraph{Limitations}
We would also like to highlight some limitations of this work that can guide future research directions.
First, our filtering pipeline relies on a single model. 
Exploring alternative architectures, contrastive training objectives, or multi-stage ranking ensembles could improve robustness to noisy or ambiguous GUI examples.
Second, while our sparse reward formulation proved effective, it remains coarse and sensitive to scaling. 
Future work should investigate more structured reward functions, including continuous measures of bounding-box accuracy, reasoning coherence, or multi-turn consistency.
Finally, the current traces are generated and optimised for brevity. 
Incorporating richer reasoning rewards on the trace-level could enhance the interpretability and depth of model reasoning.

\bibliographystyle{unsrt}  
\bibliography{references}  %%% Remove comment to use the external .bib file (using bibtex).

\appendix
\section{Data}
\subsection{Filtering Pipeline}

\paragraph{Task Difficulty}
With regards to determining challenging instances, we follow the guidelines for compute use and mobile agent from Qwen-2.5-VL\footnote{https://github.com/QwenLM/Qwen3-VL/tree/main}.
We generate one zero-shot prediction for each training examples in the original datasets of AriaUI and ShowUI. 
If the center of the predicted bounding box is not within the target region, we consider this example as challenging.

\paragraph{Diverse Training Examples}
To select diverse training examples, given the screenshot of GUI and the instruction, we perform one forward pass from Qwen-2.5-VL 3B and obtain the last hidden state from the final layer of the model.
We view this as the final representation of the multimodal sequence that captures both the image and the instruction. 
Following prior work \citep{cai2021isotropy} on clustering embeddings, we apply PCA (dim=768) on the latent representation of each example.
We then apply kmeans using 10\% of the initial data as centroids and select the example that is closest to each cluster.
We also note that we applied clustering solely on AriaUI benchmark as it contains significantly more examples compared to ShowUI.

\begin{figure*}[tb]
    \centering
    \begin{tcolorbox}[title=Prompt for Generating Chain-of-Thought Traces]

\textbf{Objective:} \\
Your task is to provide guidance on a visual grounding task, where given an image of a graphical user interface, and an instruction, the objective is to provide the correct bounding box that matches the instruction.

In the image that you are given there is a red ground truth bounding box and the instruction: \texttt{<instruction>}.

\vspace{2mm}
Below are some examples showing expected responses:

\begin{enumerate}
    \item \textbf{Example 1:}
    \begin{itemize}
        \item Instruction: `Close the current window'.
        \item Response: The image features an open Firefox browser window. To close the current window, click on `X' icon on the top-right corner.
    \end{itemize}

    \item \textbf{Example 2:}
    For each translated question-answer pair, check for errors. For example you can identify:
    \begin{itemize}
        \item Instruction: `Use the ``Insert'' button to add code'.
        \item Response: The image shows a LibreOffice document, the `Insert' button is featured on the top menu next to the `View' and `Format' options.
    \end{itemize}

    \item \textbf{Example 3:}
    \begin{itemize}
        \item Instruction: `Select Lohit Assamese font'.
        \item Response: The image displays a character formatting dialog in LibreOffice Writer. The `Lohit Assamese' font is located under the family options.
    \end{itemize}

\end{enumerate}

You need to guide the user by an explanation that allows the user to identify the red bounding box.\\

\textbf{Criteria for explanation:}
\begin{itemize}
    \item Succinct using 2 sentences at most.
    \item Avoid explicitly mentioning or referring to the highlighted red bounding box as the user needs to infer it directly from your explanation.
    \item Overall short summary of the ui.
    \item If the bounding box covers an element then provide a description tailored to the input instruction of the red groundtruth bounding box. If the bounding box covers plain text in the interface, refer to that text. DO NOT MENTION THAT THE BOUNDING BOX IS ALREADY HIGHLIGHTED.
    \item Be direct but not in the second person.
\end{itemize}

\vspace{2mm}
\textbf{Response format:}\\
You must return your response in a json format: \{"response": \texttt{<response>}\}, where \texttt{<response>} is your explanation.

\end{tcolorbox}
\caption{Guidelines generating chain-of-thought traces.}
\label{fig:cot_prompt}
\end{figure*}

\subsection{Chain-of-thought traces}\label{appendix:cot}
We extract reasoning traces in the form of explanations for each filtered instance using GPT-4o mini. 
\cref{fig:cot_prompt} shows the full prompt for extracting these traces. 
The objective of this step is to provide guidance for the fine-tuning model when deriving the solution.
During extraction, we use SOM \citep{yang2023set} to label the corresponding region and provide three in-context examples that we manually selected.
We iterated over the generation prompt as during preliminary extraction we observed that the model used the prolonged generations and often referred to the SOM directly essentially providing the solution on the image.

\begin{figure*}[tb]
\minipage{0.49\textwidth}
        \centering
        \small
        % \small
        \renewcommand{\arraystretch}{1.2}
        % \begin{adjustbox}{center}
        % \addtolength{\tabcolsep}{-0.5em}
        \begin{tabular}{@{}l c c @{}} 
            \textbf{Source} & \textbf{\# Instances} & \textbf{\# Image-Instr-Bbox Triplets}\\  
            \toprule
            AriaUI Desktop & 2668 & 29.3k\\
            AriaUI Mobile & 5035 & 52.9k\\
            AriaUI Web & 7635 & 45.7k \\
            ShowUI Desktop & 101 & 2.1k\\
            \midrule
            Total & 15k & 130k\\
            \bottomrule
        \end{tabular}
        \subcaption{Number of instances and positive image-instruction-bbox triplets per dataset source used to train the ranker model.}
        \label{tab:num_instances_ranker}
\endminipage\hfill
\minipage{0.49\textwidth}
  \includegraphics[width=\linewidth]{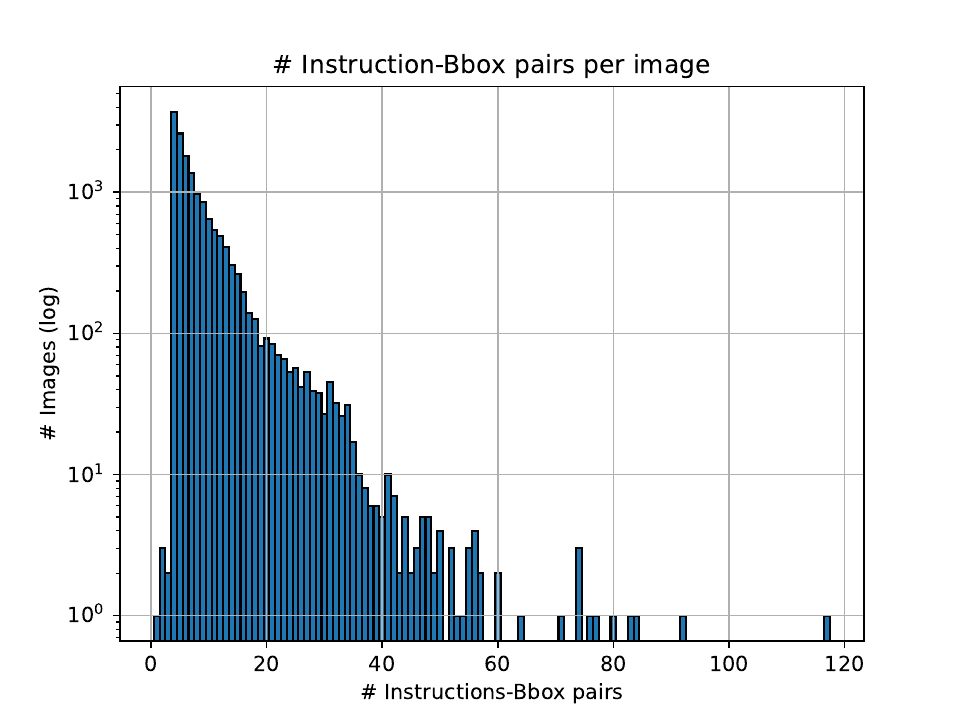}
  \subcaption{Distribution of instruction-bbox pairs.}\label{fig:histogram_ranker}
\endminipage
\end{figure*}

%https://docs.google.com/drawings/d/1VvA-ruDTzoltAgHuXi3JVMBtYN4_hv-WjsMO8f3EBYg/edit?usp=sharing
\begin{figure*}[tb]
  \includegraphics[width=\linewidth]{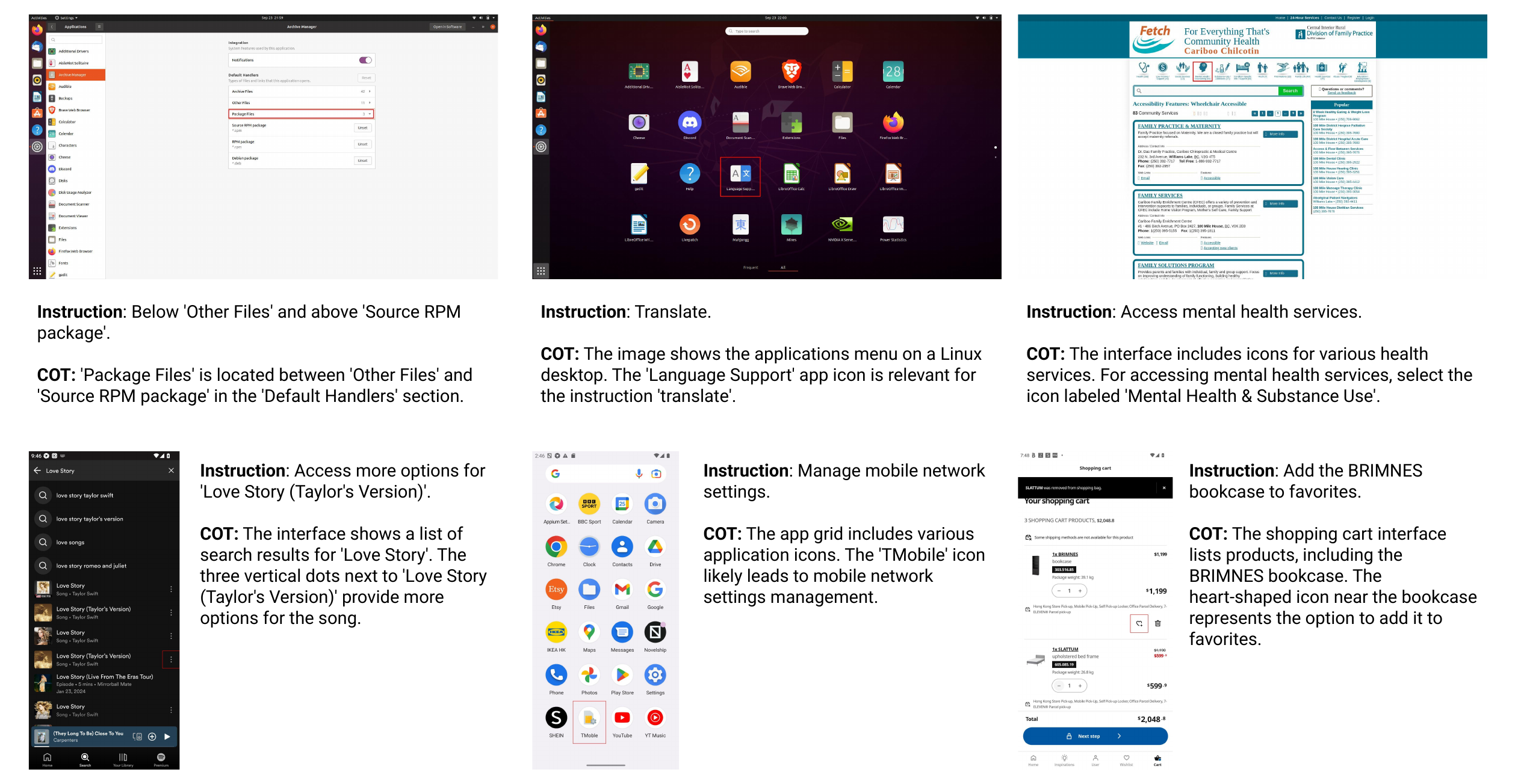}
  \caption{Examples of extracted chain-of-thought traces from desktop, web, and mobile interfaces. For clarity the correct bounding box is already drawn in the interface. Note that the instruction can target an element by describing its position, or point to an icon with underlying semantics. The goal of the trace here is to provide guidance by describing the target element or the semantics of that icon.}\label{fig:examples_cot}
\end{figure*}

%https://docs.google.com/drawings/d/16Oex1Q-p6b90alDAG0g6xkYXDaMW9lljCWnPTfGsEwU/edit?usp=sharing
\section{Experiments}\label{sec:appendix_experiments}

\paragraph{Implementation Details} \cref{tab:training-hyperparams} shows the training hyperparameters for all of our models.
Note that for a fair comparison, we use the same set of values for each training regime. 
To determine the best LoRA configuration in each experiment, we selected the model with the highest performance on ScreenSpot.
Across all settings, we observed that the lower learning rate yielded the best performance in our case.
Furthermore, we note that in the case of SFT with chain-of-thought, the length of the traces can vary significantly.
All experiments were conducted using 2$\times$ A100 80GB.

\begin{table*}[tb]
    \centering
    % \scriptsize
    \small
    \renewcommand{\arraystretch}{1.2}
        \begin{tabular}{@{}l c c c c@{}} 
            \toprule
            \textbf{Hyperparameter} & \textbf{Ranker} & \textbf{SFT} & \textbf{SFT+CoT} & \textbf{GRPO}\\
            \midrule
            global batch size & 64 & \{16, 32\} & \{16, 32\} & \{16, 32\}\\
            num epochs & 2 & \{1, 2, 4\} & \{1, 2, 4\} & \{1, 2, 4\}\\
            lr & \{1e-5, 5e-5\} & \{1e-6, 5e-6, 1e-5\} & \{1e-6, 5e-6, 1e-5\} & \{1e-6, 5e-6, 1e-5\}\\
            lr schedule & cosine decay & cosine decay & cosine decay & cosine decay\\
            lr warmup & 0.1 & 0.1 & 0.1 & 0.1\\
            number of epochs & 2 & \{1, 2, 4\} & \{1, 2, 4\} & \{1, 2, 4\}\\
            optimizer & AdamW & AdamW  & AdamW  & AdamW \\
            adam betas & (0.9, 0.999) & (0.9, 0.999) & (0.9, 0.999) & (0.9, 0.999)\\
            adam epsilon & 1e-8 & 1e-8 & 1e-8 & 1e-8\\
            weight decay & 0.1 & 0.1 & 0.1 & 0.1\\
            DeepSpeed Stage & 3 & 3 & 3 & 3\\
            \midrule
            lora modules & - & {all linear layers V/L/VL} & {all linear layers V/L/VL} & {all linear layers VL}\\
            lora rank/alpha & - & \multicolumn{3}{c}{\{r=(32, 64, 128, 256), alpha=(0.5r, 2r)\}}\\
            min pixels & 3136 & 3136 & 3136 & 3136 \\
            max pixels & \{4028160, 12845056\} & 846720 & 846720 & 846720\\
            \midrule
            num generations & - & - & - & 16\\
            temperature  & - & - & - & 1\\
            top\_p  & - & - & - & 1.0\\
            top\_k  & - & - & - & - \\
            KL beta  & - & - & - & 0\\
            reward scaling  & - & - & - & 1 \\
            \bottomrule
        \end{tabular}
    
    \caption{Hyperparameters used to train all models. For LoRA adapters V: applied on only the vision encoder and the connector, L: applied on only the language backbone, VL: applied on the entire model including the vision encoder, connector, and the language backbone.}
    \label{tab:training-hyperparams}
\end{table*}

\subsection{Ranker}\label{sec:appendix_ranker}
\cref{tab:ranker_detailed_screenspot}, \cref{tab:ranker_detailed_screenspotv2}, \cref{tab:ranker_detailed_osworld} illustrate the per-domain performance of our ranking model for ScreenSpot, ScreenSpotv2 and Osworld-G relative to 3B/7B/32B variants.

\begin{table*}[tb]
    \centering
    \scriptsize
    % \small
    \renewcommand{\arraystretch}{1.2}
    % \addtolength{\tabcolsep}{-0.3em}
        \begin{tabular}{@{}l |cccc | cccccc@{}} 
            \toprule
            \textbf{Model} & F1 & Acc & Prec & Rec & F1 (Y) & F1 (N) & Prec (Y) & Prec (N) & Rec (Y)& Rec (N)\\
            \midrule
            \multicolumn{11}{c}{ScreenSpot} \\
            \midrule
            Qwen-2.5-VL 3B$^*$ & 76.2 & 77.6 & 76.1 & 76.3 & 70.5 & 81.9 & 69.7 & 82.5 & 71.3 & 81.4 \\
            Qwen-2.5-VL 7B$^*$ & 80.3 & 82.3 & 79.1 & 83.1 & 73.9 & 86.6 & 65.4 & 92.8 & 85.0 & 81.2 \\
            Qwen-2.5-VL 3B Ranker & \textbf{93.8} & \textbf{94.2} & \textbf{93.9} & \textbf{93.8} & \textbf{92.4} & \textbf{95.3} & \textbf{92.8} & \textbf{95.0} & \textbf{92.1} & \textbf{95.5} \\
            \midrule
            \multicolumn{11}{c}{ScreenSpotv2} \\
            \midrule
            Qwen-2.5-VL 3B$^*$ & 78.0 & 78.3 & 78.0 & 78.0 & 75.5 & 80.5 & 75.6 & 80.4 & 75.4 & 80.6 \\
            Qwen-2.5-VL 7B$^*$ & 82.4 & 83.2 & 81.8 & 84.6 & 78.7 & 86.1 & 70.2 & 93.4 & 89.4 & 79.8 \\
            Qwen-2.5-VL 3B Ranker & \textbf{94.5} & \textbf{94.6} & \textbf{94.5} & \textbf{94.5} & \textbf{93.8} & \textbf{95.1} & \textbf{93.5} & \textbf{95.4} & \textbf{94.2} & \textbf{94.9}\\
            \midrule
            \multicolumn{11}{c}{Osworld-G} \\
            \midrule
            Qwen-2.5-VL 3B$^*$ & 65.7 & 74.5 & 64.5 & 69.3 & 48.2 & 83.1 & 40.0 & 89.1 & 60.9 & 77.8 \\
            Qwen-2.5-VL 7B$^*$ & 66.1 & 76.3 & 64.7 & 73.5 & 47.6 & 84.6 & 36.2 & 93.2 & 69.5 & 77.5 \\
            Qwen-2.5-VL 3B Ranker & \textbf{71.8} & \textbf{73.1} & \textbf{77.1} & \textbf{72.7} & \textbf{65.8} & \textbf{77.9} & \textbf{87.0} & \textbf{67.3} & \textbf{52.9} & \textbf{92.4}\\
            \bottomrule
        \end{tabular}
        \caption{Performance of our ranking model on ScreenSpot, Screenspotv2 and Osworld-G benchmarks. Positive and negative classes are denoted as (Y), (N) respectively. $^*$ denotes zero-shot evaluation.}
        \label{tab:ranker_overall}
\end{table*}

\begin{table*}[tb]
    \centering
    % \scriptsize
    \scriptsize
    \renewcommand{\arraystretch}{1.2}
    % \addtolength{\tabcolsep}{-0.3em}
        \begin{tabular}{@{}l |cc | cc | cc | cc | cc | cc |cc | cc@{}} 
            \toprule
            \textbf{Model} & \multicolumn{2}{c|}{\textbf{Windows}} & \multicolumn{2}{c|}{\textbf{MACOS}} & \multicolumn{2}{c|}{\textbf{IOS}} & \multicolumn{2}{c|}{\textbf{Android}} & \multicolumn{2}{c|}{\textbf{Gitlab}} & \multicolumn{2}{c|}{\textbf{Shop}} & \multicolumn{2}{c|}{\textbf{Forum}} & \multicolumn{2}{c}{\textbf{Tool}} \\
            & Text & Icon & Text & Icon & Text & Icon & Text & Icon & Text & Icon & Text & Icon & Text & Icon & Text & Icon \\
            \midrule
            Qwen-2.5-VL 3B$^*$ & 77.7 & 51.2 & 79.7 & 66.5 & 86.2 & 73.0 & 82.1 & 63.2 & 81.9 & 57.5 & 85.9 & 71.0 & 87.3 & 73.8 & 73.9 & 75.2 \\
            Qwen-2.5-VL 7B$^*$ & 81.3 & 60.5 & 81.9 & 69.4 & 89.8 & 78.5 & 82.1 & 61.0 & 88.5 & 66.8 & 82.6 & 79.5 & 88.2 & 79.7 & 85.2 & 83.2 \\
            Qwen-2.5-VL 32B$^*$ & 94.7 & 70.7 & 96.6 & 80.9 & 94.3 & 86.5 & 85.7 & 71.1 & 91.7 & 81.6 & 89.4 & 86.0 & 88.2 & 87.5 & 90.1 & 89.3 \\
            \midrule
            Qwen-2.5-VL 3B Ranker & 96.4 & 80.6 & 98.7 & 86.1 & 98.9 & 93.6 & 96.6 & 89.5 & 97.5 & 82.9 & 97.8 & 93.5 & 93.6 & 92.8 & 89.7 & 93.5 \\
            
            \bottomrule
        \end{tabular}
        \caption{Ranking performance on ScreenSpot. $^*$ denotes zero-shot evaluation.}
        \label{tab:ranker_detailed_screenspot}
\end{table*}

\begin{table*}[tb]
    \centering
    % \scriptsize
    \scriptsize
    \renewcommand{\arraystretch}{1.2}
    % \addtolength{\tabcolsep}{-0.3em}
        \begin{tabular}{@{}l |cc | cc | cc | cc | cc | cc |cc | cc@{}} 
            \toprule
            \textbf{Model} & \multicolumn{2}{c|}{\textbf{Windows}} & \multicolumn{2}{c|}{\textbf{MACOS}} & \multicolumn{2}{c|}{\textbf{IOS}} & \multicolumn{2}{c|}{\textbf{Android}} & \multicolumn{2}{c|}{\textbf{Gitlab}} & \multicolumn{2}{c|}{\textbf{Shop}} & \multicolumn{2}{c|}{\textbf{Forum}} & \multicolumn{2}{c}{\textbf{Tool}} \\
            & Text & Icon & Text & Icon & Text & Icon & Text & Icon & Text & Icon & Text & Icon & Text & Icon & Text & Icon \\
            \midrule
            Qwen-2.5-VL 3B$^*$ & 79.8 & 51.9 & 79.7 & 67.1 & 86.2 & 72.6 & 81.9 & 70.4 & 82.8 & 61.5 & 83.8 & 69.8 & 84.6 & 74.9 & 81.4 & 74.6 \\
            Qwen-2.5-VL 7B$^*$ & 85.0 & 62.3 & 83.7 & 68.7 & 89.7 & 81.0 & 82.3 & 66.2 & 88.0 & 65.8 & 84.6 & 78.7 & 88.4 & 78.6 & 90.0 & 86.6 \\
            \midrule
            Qwen-2.5-VL 3B Ranker & 94.9 & 81.7 & 99.5 & 87.1 & 99.1 & 92.2 & 99.3 & 92.6 & 97.0 & 81.4 & 96.6 & 91.4 & 92.3 & 88.7 & 97.5 & 88.4 \\
            \bottomrule
        \end{tabular}
        \caption{Ranking performance on ScreenSpotv2. $^*$ denotes zero-shot evaluation.}
        \label{tab:ranker_detailed_screenspotv2}
\end{table*}

\begin{table*}[tb]
    \centering
    % \scriptsize
    \scriptsize
    \renewcommand{\arraystretch}{1.2}
    % \addtolength{\tabcolsep}{-0.3em}
    \begin{tabular}{@{}l |cc cc cc cc @{}} 
        \textbf{Model} & \textbf{Text Matching} & \textbf{Element Recognition} & \textbf{Layout Understanding} & \textbf{Finegrained Manipulation} \\
        \toprule
        Qwen-2.5-VL 3B$^*$ & 70.5 & 63.1 & 66.5 & 60.5 \\
        Qwen-2.5-VL 7B$^*$ & 69.9 & 64.6 & 67.8 & 58.0 \\
        \midrule
        Qwen-2.5-VL 3B Ranker &  85.2 & 63.9 & 66.8 & 79.0 \\
        \bottomrule
    \end{tabular}
    \caption{Ranking performance on Osworld-G. $^*$ denotes zero-shot evaluation.}
    \label{tab:ranker_detailed_osworld}
\end{table*}

\subsection{Data Ablations}
\cref{tab:screenspot_data_ablations_full}, and \cref{tab:screenspotv2_data_ablations_full} illustrates detailed results regarding our data ablations. 
We report the grounding accuracy on ScreenSpot and Screenspotv2 with and without data filtering across all lora configurations.

\begin{table*}[tb]
    \centering
    % \scriptsize
    \scriptsize
    \renewcommand{\arraystretch}{1.2}
    \addtolength{\tabcolsep}{-0.1em}
        \begin{tabular}{@{}l c|cc | cc | cc | cc | cc | cc |cc | cc | c@{}} 
            \toprule
            \textbf{Params} & \textbf{Filtering} & \multicolumn{2}{c|}{\textbf{Windows}} & \multicolumn{2}{c|}{\textbf{MACOS}} & \multicolumn{2}{c|}{\textbf{IOS}} & \multicolumn{2}{c|}{\textbf{Android}} & \multicolumn{2}{c|}{\textbf{Gitlab}} & \multicolumn{2}{c|}{\textbf{Shop}} & \multicolumn{2}{c|}{\textbf{Forum}} & \multicolumn{2}{c|}{\textbf{Tool}} & \textbf{Micro}\\
            (r/a) &  & Text & Icon & Text & Icon & Text & Icon & Text & Icon & Text & Icon & Text & Icon & Text & Icon & Text & Icon & \textbf{Avg} \\
            \midrule
            32/16 & \xmark & 64.1 & 86.7 & 56.6 & 86.5 & 62.6 & 93.2 & 68.9 & 91.2 & 54.1 & 86.5 & 70.0 & 84.8 & 72.1 & 71.7 & 60.6 & 79.5 & 76.6 \\
            32/16 & \cmark & 70.3 & 91.8 & 61.8 & 94.8 & 70.1 & 98.7 & 68.0 & 92.8 & 62.2 & 92.3 & 76.7 & 88.1 & 67.4 & 82.6 & 65.2 & 79.5 & 81.0 \\
            \midrule
            64/32 & \xmark & 57.8 & 90.8 & 55.3 & 87.5 & 62.6 & 93.9 & 72.1 & 88.8 & 51.4 & 86.5 & 68.3 & 84.8 & 72.1 & 71.7 & 63.6 & 74.0 & 76.4 \\
            64/32 & \cmark & 73.4 & 92.9 & 63.2 & 91.7 & 72.0 & 98.0 & 69.7 & 91.2 & 59.5 & 92.3 & 70.0 & 86.4 & 65.1 & 84.8 & 63.6 & 75.3 & 80.4 \\
            \midrule
            128/64 & \xmark & 53.1 & 87.8 & 52.6 & 92.7 & 71.0 & 92.6 & 68.9 & 88.0 & 56.8 & 90.4 & 66.7 & 86.4 & 74.4 & 76.1 & 71.2 & 74.0 & 77.3 \\
            128/64 & \cmark & 68.8 & 91.8 & 63.2 & 92.7 & 70.1 & 98.7 & 68.9 & 95.2 & 56.8 & 92.3 & 76.7 & 88.1 & 69.8 & 80.4 & 60.6 & 75.3 & 80.5 \\
            \midrule
            256/128 & \xmark & 39.1 & 85.7 & 52.6 & 87.5 & 71.0 & 94.6 & 68.0 & 94.4 & 54.1 & 90.4 & 76.7 & 84.8 & 76.7 & 80.4 & 71.2 & 72.6 & 77.3 \\
            256/128 & \cmark & 75.0 & 94.9 & 60.5 & 94.8 & 69.2 & 98.7 & 65.6 & 93.6 & 54.1 & 90.4 & 76.7 & 86.4 & 62.8 & 84.8 & 63.6 & 72.6 & 80.2 \\
            \midrule
            32/64 & \xmark & 57.8 & 87.8 & 52.6 & 90.6 & 69.2 & 93.2 & 70.5 & 92.0 & 56.8 & 88.5 & 78.3 & 88.1 & 81.4 & 78.3 & 66.7 & 76.7 & 78.6 \\
            32/64 & \cmark & 71.9 & 93.9 & 64.5 & 92.7 & 73.8 & 98.0 & 66.4 & 92.8 & 67.6 & 94.2 & 78.3 & 89.8 & 69.8 & 82.6 & 59.1 & 78.1 & 81.4 \\
            \midrule
            64/128 & \xmark & 45.3 & 86.7 & 50.0 & 89.6 & 73.8 & 91.2 & 68.9 & 92.0 & 62.2 & 84.6 & 80.0 & 86.4 & 79.1 & 76.1 & 69.7 & 78.1 & 77.8 \\
            64/128 & \cmark & 70.3 & 90.8 & 55.3 & 92.7 & 68.2 & 96.0 & 67.2 & 91.2 & 54.1 & 88.5 & 85.0 & 91.5 & 67.4 & 82.6 & 56.1 & 72.6 & 78.9 \\
            \midrule
            128/256 & \xmark & 50.0 & 81.6 & 50.0 & 87.5 & 64.5 & 92.6 & 70.5 & 88.8 & 54.1 & 90.4 & 75.0 & 88.1 & 83.7 & 84.8 & 72.7 & 78.1 & 77.1 \\
            128/256 & \cmark & 62.5 & 89.8 & 51.3 & 94.8 & 63.6 & 93.9 & 68.9 & 92.0 & 46.0 & 94.2 & 78.3 & 89.8 & 60.5 & 82.6 & 63.6 & 74.0 & 77.8 \\
            \midrule
            256/512 & \xmark & 53.1 & 76.5 & 34.2 & 81.3 & 53.3 & 91.9 & 59.8 & 81.6 & 46.0 & 84.6 & 66.7 & 83.1 & 67.4 & 78.3 & 54.6 & 69.9 & 69.4 \\
            256/512 & \cmark & 40.6 & 85.7 & 36.8 & 88.5 & 53.3 & 91.9 & 59.8 & 86.4 & 40.5 & 80.8 & 65.0 & 84.8 & 48.8 & 82.6 & 53.0 & 68.5 & 69.7 \\
            \bottomrule
        \end{tabular}
        \caption{Grounding accuracy on ScreenSpot with and without data filtering. LoRA adapters are introduced within all linear layers of the model. Across all configurations, the data filtering approach yields greater performance (VL).}
        \label{tab:screenspot_data_ablations_full}
\end{table*}

\begin{table*}[tb]
    \centering
    % \scriptsize
    \scriptsize
    \renewcommand{\arraystretch}{1.2}
    \addtolength{\tabcolsep}{-0.1em}
        \begin{tabular}{@{}l c|cc | cc | cc | cc | cc | cc |cc | cc | c@{}} 
            \toprule
            \textbf{Params} & \textbf{Filtering} & \multicolumn{2}{c|}{\textbf{Windows}} & \multicolumn{2}{c|}{\textbf{MACOS}} & \multicolumn{2}{c|}{\textbf{IOS}} & \multicolumn{2}{c|}{\textbf{Android}} & \multicolumn{2}{c|}{\textbf{Gitlab}} & \multicolumn{2}{c|}{\textbf{Shop}} & \multicolumn{2}{c|}{\textbf{Forum}} & \multicolumn{2}{c|}{\textbf{Tool}} & \textbf{Micro}\\
            (r/a) &  & Text & Icon & Text & Icon & Text & Icon & Text & Icon & Text & Icon & Text & Icon & Text & Icon & Text & Icon & \textbf{Avg} \\
            \midrule
            32/16 & \xmark & 66.7 & 87.5 & 68.3 & 91.7 & 70.0 & 93.5 & 73.6 & 92.5 & 58.6 & 90.9 & 63.5 & 88.5 & 71.8 & 77.5 & 59.3 & 80.3 & 79.6 \\
            32/16 & \cmark & 74.6 & 91.7 & 68.3 & 98.8 & 78.0 & 97.8 & 71.4 & 94.2 & 58.6 & 93.2 & 67.0 & 86.2 & 69.2 & 80.0 & 68.5 & 84.9 & 82.6 \\
            \midrule
            64/32 & \xmark & 58.7 & 92.7 & 66.7 & 94.1 & 69.0 & 92.8 & 74.7 & 92.5 & 58.6 & 84.1 & 62.6 & 87.7 & 69.2 & 72.5 & 66.7 & 84.9 & 79.5 \\
            64/32 & \cmark & 76.2 & 94.8 & 69.8 & 96.4 & 73.0 & 97.1 & 79.1 & 92.5 & 62.1 & 93.2 & 68.7 & 86.9 & 66.7 & 80.0 & 64.8 & 81.8 & 82.7 \\
            \midrule
            128/64 & \xmark & 58.7 & 89.6 & 57.1 & 97.6 & 76.0 & 94.2 & 75.8 & 92.5 & 69.0 & 88.6 & 60.0 & 88.5 & 74.4 & 77.5 & 66.7 & 78.8 & 80.0 \\
            128/64 & \cmark & 69.8 & 93.8 & 69.8 & 96.4 & 76.0 & 98.6 & 78.0 & 95.0 & 55.2 & 90.9 & 71.3 & 88.5 & 71.8 & 77.5 & 61.1 & 80.3 & 82.9 \\
            \midrule
            256/128 & \xmark & 41.3 & 84.4 & 57.1 & 92.9 & 76.0 & 94.9 & 78.0 & 95.8 & 58.6 & 88.6 & 71.3 & 88.5 & 74.4 & 77.5 & 64.8 & 77.3 & 79.6 \\
            256/128 & \cmark & 76.2 & 95.8 & 63.5 & 98.8 & 74.0 & 97.1 & 74.7 & 94.2 & 62.1 & 90.9 & 73.9 & 88.5 & 69.2 & 80.0 & 64.8 & 77.3 & 82.9 \\
            \midrule
            32/64 & \xmark & 58.7 & 89.6 & 58.7 & 95.2 & 71.0 & 93.5 & 78.0 & 93.3 & 69.0 & 88.6 & 68.7 & 87.7 & 79.5 & 77.5 & 59.3 & 83.3 & 80.5 \\
            32/64 & \cmark & 73.0 & 94.8 & 74.6 & 97.6 & 78.0 & 98.6 & 69.2 & 95.0 & 65.5 & 93.2 & 73.0 & 88.5 & 71.8 & 77.5 & 57.4 & 78.8 & 83.2 \\
            \midrule
            64/128 & \xmark & 47.6 & 89.6 & 54.0 & 95.2 & 77.0 & 92.0 & 78.0 & 95.0 & 62.1 & 86.4 & 71.3 & 91.5 & 76.9 & 75.0 & 70.4 & 78.8 & 80.7 \\
            64/128 & \cmark & 71.4 & 93.8 & 58.7 & 96.4 & 73.0 & 96.4 & 78.0 & 92.5 & 55.2 & 88.6 & 71.3 & 90.8 & 69.2 & 80.0 & 61.1 & 72.7 & 81.5 \\
            \midrule
            128/256 & \xmark & 49.2 & 82.3 & 55.6 & 92.9 & 67.0 & 92.0 & 74.7 & 91.7 & 62.1 & 86.4 & 67.0 & 86.9 & 79.5 & 85.0 & 74.1 & 78.8 & 78.5 \\
            128/256 & \cmark & 66.7 & 91.7 & 55.6 & 97.6 & 70.0 & 94.2 & 72.5 & 95.8 & 48.3 & 93.2 & 64.4 & 90.0 & 64.1 & 80.0 & 57.4 & 83.3 & 80.0 \\
            \midrule
            256/512 & \xmark & 54.0 & 78.1 & 36.5 & 86.9 & 58.0 & 92.0 & 63.7 & 84.2 & 51.7 & 84.1 & 56.5 & 80.8 & 66.7 & 80.0 & 46.3 & 77.3 & 71.2 \\
            256/512 & \cmark & 41.3 & 86.5 & 41.3 & 91.7 & 60.0 & 92.0 & 64.8 & 90.0 & 41.4 & 81.8 & 59.1 & 87.7 & 51.3 & 75.0 & 50.0 & 75.8 & 72.6 \\
            \bottomrule
        \end{tabular}
        \caption{Grounding accuracy on ScreenSpotv2 with and without data filtering. LoRA adapters are introduced within all linear layers of the model. Across all configurations, the data filtering approach yields greater performance (VL).}
        \label{tab:screenspotv2_data_ablations_full}
\end{table*}

\end{document}